\RequirePackage{fix-cm}
\documentclass[twocolumn]{svjour3}       
\smartqed  
\usepackage{graphicx}
\usepackage{color}
\usepackage{textcomp}
\usepackage{amsmath}
\usepackage{hyperref}
\usepackage{hyperref}
\usepackage{chngcntr}
\usepackage[toc,page]{appendix}
\usepackage{cite}

\usepackage{etoolbox}
\makeatletter
\patchcmd\@combinedblfloats{\box\@outputbox}{\unvbox\@outputbox}{}{%
   \errmessage{\noexpand\@combinedblfloats could not be patched}%
}%
 \makeatother
\begin{document}
\emergencystretch 3em

\title{A Predictive Account of Caf\'e Wall Illusions \\Using a Quantitative Model
}


\author{Nasim Nematzadeh \and
        David M.W. Powers 
}

\institute{N. Nematzadeh \at
              College of Science and Engineering, Flinders University, Adelaide, Australia \\
              Faculty of Mechatronics, Dept. of Science and Engineering, Karaj Branch, Islamic Azad University (KIAU), Karaj-Alborz, Iran \\
              Tel.: +61-08-74219555\\
              Fax: +61-08-82013399\\
              \email{nasim.nematzadeh@flinders.edu.au}           
           \and
           D. M. W. Powers \at
              College of Science and Engineering, Flinders University, Adelaide, Australia
}
\maketitle
\begin{abstract}
This paper explores the tilt illusion effect in the Caf\'e Wall pattern using a classical Gaussian Receptive Field model.  In this illusion, the mortar lines are misperceived as diverging or converging rather than horizontal. We examine the capability of a simple bioplausible filtering model to recognize different degrees of tilt in the Caf\'e Wall illusion based on different characteristics of the pattern. Our study employed a Difference of Gaussians model of retinal to cortical ``ON'' center and/or ``OFF'' center receptive fields. A wide range of parameters of the stimulus, for example mortar width, luminance, tiles contrast, phase of the tile displacement, have been studied for their effects on the induced tilt in the Caf\'e Wall illusion. Our model constructs an edge map representation at multiple scales that reveals tilt cues and clues involved in the illusory perception of the Caf\'e Wall pattern. We show here that our model can not only detect the tilt in this pattern, but also allows us to predict the strength of the illusion and quantify the degree of tilt. For the first time, quantitative predictions of a model are reported for this stimulus considering different characteristics of the pattern. The results of our simulations are consistent with previous psychophysical findings across the full range of Caf\'e Wall variations tested. Our results also suggest that the Difference of Gaussians mechanism is at the heart of the effects explained by, and the mechanisms proposed for, the Irradiation, Brightness Induction, and Bandpass Filtering models.\\
\end{abstract}
\section{Introduction}
\label{intro}
\subsection{Illusions and their explanations}
\label{1.1}
Visual illusions have the potential to give insight into the biology of vision \cite{Changizi08,Eagleman01,Grossberg88}. They further open windows for solving the engineering and application specific problems relevant to image processing tasks including edge detection and feature selection, as we seek to attain human level performance. Many illusions have been used to evaluate such effects, notably the Caf\'e Wall \cite{Gregory79,McCourt83,Morgan86} and Mach Bands \cite{Kingdom92,Pessoa96}.
The Caf\'e Wall is a member of the Twisted Cord family of Geometrical Illusions, for which a number of explanations have been proposed \cite{Gregory79,Earle93,Woodhouse87}. The Munsterberg version of this pattern is a chessboard with dark very thin separators between shifted rows of black and white tiles, giving the illusion of tilted contours dividing the rows. 

\begin{figure*}
	\centering
	\includegraphics[width=0.75\textwidth]{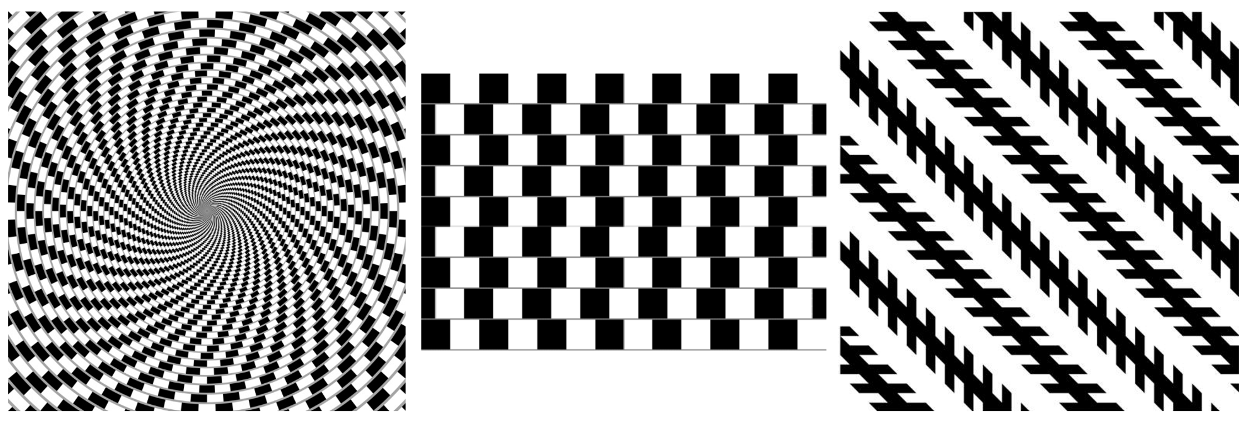}
	\caption{Left: The Spiral Caf\'e Wall illusion \cite{Kitaoka07}, Center: The Caf\'e Wall pattern, Right: The Zollner illusion \cite{ZollnerWeb16}.}
	\label{fig:1}   
\end{figure*}

The Caf\'e Wall pattern (\hyperref[fig:1]{Fig. \ref*{fig:1}}-center) has grey stripes interpreted as mortar lines dividing shifted rows of black and white tiles, inducing a perception of diverging and converging of the mortar lines. Morgan and Moulden \cite{Morgan86} suggest the mortar lines are critical for the strength of illusion and that the illusion has its highest strength when the luminance of the mortar is intermediate relative to the luminance of the tiles. We consider the whole range of luminance for mortar lines from Black to White as different variations of the Caf\'e Wall illusion. Other significant variations include the Hollow Square \cite{Woodhouse87,Bressan85}, Fraser Pattern \cite{Fraser08}, Spiral Caf\'e Wall (\hyperref[fig:1]{Fig. \ref*{fig:1}}-left), Spiral Fraser; and even variations of the Zollner Illusion (\hyperref[fig:1]{Fig. \ref*{fig:1}}-right) \cite{McCourt83,Bressan85,Westheimer07} where small parallel line segments (inducing horizontal/vertical lines) on parallel lines result in perceiving these lines as being convergent or divergent. It has been suggested that the Caf\'e Wall illusion originates from the inducing effect of Twisted Cord \cite{Fraser08} elements and then integration of these to an extended continuous contour along the whole mortar line \cite{Grossberg85,Moulden79}.

Over the last few decades, many low-level and high-level models have been proposed with the aim of explaining various Geometrical/Tilt Illusions. However, there are many issues unaddressed across the range of the Geometrical Illusions and their underlying percepts. In particular, modeling the illusion effect as it is perceived is a challenging task in Computer Vision models. Some of the explanations for the Caf\'e Wall illusion are based on ``high-level'' models \cite{Gregory79,Kitaoka07}, but many rely on ``low-level bioplausible'' models of simple filtering techniques \cite{Morgan86,Earle93}. 

One of the first explanations for the Caf\'e Wall illusion was the \emph{Irradiation Hypothesis} \cite{Moulden79}. This hypothesis was first introduced by Helmholtz \cite{Helmholtz11}, proposing that a compressive transform causes a shift of high contrast boundary towards the dark side of its inflection \cite{Morgan86}. The limitation of the Irradiation Hypothesis as a sole explanation of the Caf\'e Wall illusion is that it does not explain why the illusion is enhanced by the mortar lines and why the pattern does not need to be of a very high contrast as explicitly required by the Irradiation Theory. So this explanation is incomplete on its own.

\emph{Border Locking} is another hypothesis used in high-level explanations for Tilt Illusions \cite{Gregory79}. The appearance of small wedges with local asymmetry is attributed to the luminance contrast of the light/dark half tiles and their integration along the rows, so that they form long wedges. Gregory and Heard \cite{Gregory79} state that the effect depends critically on luminance and that it disappears if the mortar line is either brighter than the light tiles, or dimmer than the dark ones.

\emph{Brightness Induction} (BI) describes a change in the direction of perceived brightness, and is another explanation for the Caf\'e Wall illusion. The change of direction can be towards the surrounds (Brightness Assimilation) or away from it (Brightness Contrast) \cite{Penacchio13}. McCourt \cite{McCourt83} has shown that a periodic field of a sine or square wave varying brightness would result in inducing brightness on a homogeneous narrow grey field positioned on top of it. He generated a new version of the Caf\'e Wall based on replacing mortar lines with patches of light, dark and grey and explained the tilt effect as Brightness Induction, noting that alternately-reversing diagonal line elements can produce the characteristic of illusory convergence in the whole image.

Fraser \cite{Fraser08} proposed another hypothesis, connecting the \emph{Twisted Cord} to the Caf\'e Wall illusion without using filtering but his hypothesis is not a complete explanation of the illusion. His idea was based on the fact that the tilt is perceived when two identically coloured tiles joined by the mortar lines at their opposite corners create a Twisted Cord element. There are some other proposed explanations such as \emph{Fourier-based Models}, but it seems that the effect arises from small local wedge shapes rather than global features. Therefore, what is needed is local frequency analysis rather than the global transformation \cite{Morgan86}.

\emph{Bandpass Spatial Filtering} is said to be performed by the retinal ganglion cells of ON-center and OFF-center receptive fields. Morgan and Moulden \cite{Morgan86} modelled this by a simple discrete Laplacian filter mask viewed as an approximation to Difference of Gaussians (DoG, noting that DoG gives the same results), and Earle and Maskell \cite{Earle93} by a Difference of Gaussians viewed as ``a very close approximation'' to a Laplacian of Gaussian (LoG).

There are some more recent explanations for the Caf\'e Wall illusion. For example the \emph{Irradiation Effect} involving enlargement of bright regions to encroach on their dark sides was identified for the shifted chessboard pattern by Westheimer \cite{Westheimer07}, who proposed a model including a combination of \emph{`light spread', `compressive nonlinearity', `center surround organization' and `border locations'}, to explain the Caf\'e Wall illusion. For retinal spatial distribution of excitation, he first calculated the convolution of image with the retinal point spread function, which then passed through a certain nonlinearity equation (Naka-Rushton), and then to the spatial center-surround transformation with the excitatory and inhibitory zones to get the final result.

A similar psychological approach to that of Gregory and Heard \cite{Gregory79} was proposed by Kitaoka as the \emph{Phenomenal Model} \cite{Kitaoka07,Kitaoka04}, which is an elemental composition approach to explain a wide variety of Tilt Illusions. Their psychological approach is based on the ``contrast polarities of a solid square and its adjacent line segment'' to explain the Caf\'e Wall illusion. They have suggested that when dark/light squares are adjacent to dark/light line segments, the direction of tilt is to counteract the square angle, otherwise the square angle is going to expand.

Ferm{\"u}ller and Malm \cite{Fermuller04} proposed a \emph{Point and Line Estimation} Model for finding the displacement of points and lines as the elementary units in images and they used their techniques to explain certain Geometrical Illusions. As applied to the Caf\'e Wall, this theory is based on categorizing the edges by saying that if mortar lines border both dark tile and bright tile, then the two edges move towards each other. On the other hand, mortar lines between two bright regions or two dark regions cause the edges to move away from each other. Their explanation is similar to Gregory and Heard \cite{Gregory79} and Kitaoka et al. \cite{Kitaoka04}.

Arai \cite{Arai05} proposed a computational nonlinear model for \emph{Contrast Enhancement} in early visual processing based on discrete maximal overlap bi-orthogonal wavelets. Arai \cite{Arai05} investigated the output of their system on Brightness/Lightness Illusions such as the Hermann Grid, the Chevreul, Mach Bands, the Todorovic, and the Caf\'e Wall.

As noted before, an important class of low-level models are motivated by the \emph{retinal ganglion cell (RGC) responses}, including the \emph{Bandpass Spatial Filtering} proposed by Morgan and Moulden \cite{Morgan86}. These retinal cells can be modeled by Differences of Gaussian (DoG) filters as an effective filtering technique for identifying the edges in the early stages of visual processing \cite{Enroth-Cugell66,Rodieck65}. The model we have used presents the simple cell responses using an \emph{edge map representation at multiple scales} derived from the Differences of Gaussians \cite{Nematzadeh15,Nematzadeh16a,Nematzadeh16b,Nematzadeh16c,Nematzadeh17a,Nematzadeh17b,Nematzadeh18}.  
We show here that the model can not only detect the tilts in variations of the Caf\'e Wall tested, but also allows us to simply and correctly predict the strength of the illusion and quantify the degree of tilt in these variations. The `predicted tilt results' are explained in the experimental result section and are consistent with previous psychophysical findings across a full range of Caf\'e Walls tested.

Elimination of the tilt effect in the Caf\'e Wall  illusion is possible through enlargement of the black tiles \cite[Fig.11]{Westheimer07}. This results in compensating for the border shifts and corner distortions. Similar elimination techniques can be used with additional superimposed dots positioned in the corners (black/white dots on white/black tiles), eliminating the corner effect \cite[Fig.5]{Fermuller04} and \cite[Figs 5C,5D]{Kitaoka04}. The elimination of the illusion is also possible by replacing the black and white tiles with equiluminant but highly contrasting colour tiles \cite{McCourt83,Westheimer08}.

There are approaches that connect `Brightness Induction' illusions and `Geometrical' illusions. The Caf\'e Wall pattern and its variations are accounted `second-order' tilt patterns \cite{Nematzadeh15,Nematzadeh17b} involving `Brightness Assimilation and Contrast' \cite{Jameson89} as well as `Border shifts' \cite{Morgan86,Westheimer07}. In a thorough review of lightness, brightness, and transparency (LBT) Kingdom \cite{Kingdom11} has noted that one of the most promising approaches for modelling brightness coding is multiscale filtering \cite{Blakeslee99} in conjunction with contrast normalization. Lateral inhibition is a common feature of most of the above models, including in particular \emph{Irradiation} \cite{Moulden79}, \emph{Brightness Induction} \cite{McCourt83}, and \emph{Bandpass filtering} \cite{Morgan86}.
\subsection{Multiscale Representation in Vision}
\label{sec:1.2}
Psychophysical and physiological findings have suggested a multiscale model of processing in the mammalian visual cortex as well as early stage processing within the retina \cite{Field07,Gauthier09,Gollisch10}. Kuffler \cite{Kuffler52} was the pioneer who recorded the activity of the retinal ganglion cells (RGCs) that exit as the optic nerve, and found two types of center surround cells. Hubel and Wiesel \cite{Hubel79,Hubel62} performed many pioneering experiments that increased our understanding of cortical visual processing.  Daugman \cite{Daugman80} showed an approximation of these cortical cells by using Gaussian windows modulated by a sinusoidal wave for this impulse response. A specific spatial orientation tuning of these cells arises from dilation of modulated Gaussian-Gabor functions \cite{Lindeberg13}.

Pyramidal image representations and scale invariant transforms are used in Computer Vision (CV) applications \cite{Marr80,Burt83,Marr82,Watt91,Rosenfeld71,Mallat96,Lowe99} and are well matched to human visual encoding. A scale-space analysis is an emergent result of image decomposition by finding the differences between pairs of scaled filters with different parameterizations, notably the Laplacian or Difference of Gaussian filters (LoG/DoG) \cite{Lindeberg13,Jacques11,Lindeberg11}, giving rise to Ricker and Marr wavelets.

Note further that self-organization models of repeated patterns of edge detectors at particular angles are well-established \cite{Malsburg73}. Higher-level spatial aggregation of regularly spaced spots or edges in turn automatically gives rise to analogues of DCT and DWT type bases, the latter with localization determined by the higher-level lateral interaction functions or the constraints of an underlying probabilistic connectivity model \cite{Powers83}.

In our visual system, light is focused into receptors which transduce it to neural impulses that are further processed in the retina \cite{Smith03}. Then the middle layer of the retina, which is the focus of our study, enhances the neural signals through the process of lateral inhibition \cite{Ratliff69}, causing an activated nerve cell in the middle layer to decrease the ability of its nearby neighbors to become active. This biological convolution with its specific `Point Spread Function' (PSF) enables feature/texture encoding of the visual scene but also, we show, leads to optical illusions.

The first layer of the retina has a nonlinear mechanism for retinal gain control, flattening the illumination component, and making it possible for the eye to see under poor light conditions \cite{Smith03}. The lateral inhibition in the middle layer of the retina thus evidences both a bandpass filtering property and an edge enhancement capability. In the final layer, we find ganglion cells whose axons exit the eye and carry the visual signals to the cortex (\emph{inter alia}).

The contrast sensitivity of the retinal ganglion cells can be modeled based on Classical Receptive Fields (CRFs), involving circular center and surround antagonism. To reveal the edge information it uses the differences and second differences of Gaussians \cite{Enroth-Cugell66,Rodieck65} or Laplacian of Gaussian (LoG) \cite{Ghosh07}. Marr and Hildreth \cite{Marr80} proposed an approximation of LoG with DoG based on a specified ratio of the standard deviations ($\sigma$) of the Gaussians \cite{Enroth-Cugell66,Rodieck65}. 

Symmetrical DoGs at multiple scales have been used for spatial filtering in our study to generate edge maps, modelling ON-center OFF-surround receptive fields (RFs). We show the model can predict the illusory tilt in the Caf\'e Wall patterns by providing both qualitative and quantitative results.

In \hyperref[sec:2]{Section \ref*{sec:2}} we describe the characteristics of a simple classical model based on \emph{Difference and Laplacian of Gaussians} (DoG, LoG), and utilize this bioplausible model to explain the Caf\'e Wall illusion qualitatively and quantitatively (\hyperref[sec:2.1]{Sections \ref*{sec:2.1}} and \hyperref[sec:2.2]{\ref*{sec:2.2}}) followed by the details of patterns investigated (\hyperref[sec:2.3]{Section \ref*{sec:2.3}}). The experimental results on variations of the Caf\'e Wall illusion are presented in \hyperref[sec:3]{Section \ref*{sec:3}} along with a summary diagram in \hyperref[fig:8]{Fig. \ref*{fig:8}} based on the strength of illusion. This is the first time a quantitative tilt predictions of a model is reported for variations of the Caf\'e Wall illusions. We conclude by highlighting the advantages and disadvantages of the model and proceed to outline a roadmap of our ongoing and future work (\hyperref[sec:4]{Section \ref*{sec:4}}). 
\section{Material and Method}
\label{sec:2}
We hypothesize that visual perception of a scene starts by extracting the multiscale edge map, and propose that a bioplausible implementation of a contrast sensitivity of retinal RFs using DoG filtering produces a stack of multiscale outputs \cite{Romeny08}. One of the first models for foveal retinal vision was proposed by Lindeberg and Florack \cite{Lindeberg94} and our model is in most respects inspired by it. Their model is based on simultaneous sampling of the image at all scales, and the edge map in our model is generated in a similar way. There are also possibilities of involvement of higher order derivatives of Gaussians, which can be seen in retinal to cortical visual processing models such as \cite{Marr80,Ghosh07,Lourens95,Young87} but there is no biological evidence for them.

Our model has some similarities with `Brightness Assimilation and Contrast' theory of early models developed by Jameson and Hurvich \cite{Jameson89} based on DoG filters with multiple spatial scales. They noted the existence of parallel processes that occur in our visual system as the result of the simultaneous appearance of sharp edges and mixed colours that define delimited regions.  They proposed that a contrast effect happens when the stimulus components are relatively large in size compared to the center of the filter, and an assimilation effect happens when components of the stimulus are small compared to the filter center.

Recent physiological findings on retinal ganglion cells (RGCs) have dramatically extended our understanding of retinal processing. Previously, it was believed that retinal lateral inhibition could not be a major contributor to the Caf\'e Wall illusion because the effect is highly directional and arises from both orientation as well as spatial frequency. Neuro-computational eye models \cite{Lindeberg13,Romeny08,Lindeberg94,Young87} have been proposed based on biological findings by considering the size variation of RGCs due to variations of the eccentricity and dendritic field size \cite{Shapley86}. It was also shown that, some RGCs have been found to have orientation selectivity similar to the cortical cells \cite{Barlow63,Weng05}. Also there is evidence for the existence of other types of retinal cells such as horizontal and amacrine cells which have elongated surround beyond the CRF size. This has lead to orientation selectivity for modelling of the eye, as retinal non-CRFs (nCRFs) implementation \cite{Carandini04,Cavanaugh02,Wei11}.

The diversity of intra-retinal circuits and different types of RGCs \cite{Field07,Gauthier09} are the biological evidence for the underlying mechanisms of retinal multiscale processing of visual data at fine to coarse scales. Also there are variations of the size of each individual RGC due to the retinal eccentricity which is the distance from the fovea \cite{Lourens95}. This indicates a high likelihood of the involvement of retinal/cortical simple cells and their early stages of processing in revealing the tilt cues inside Tile Illusion patterns and in the Caf\'e Wall in particular.

A complete vision theory, modelling the first stages of visual processing, has been presented by Lindeberg \cite{Lindeberg13}. This  reflects the properties of the receptive field (RF) profiles, in particular, `scale covariance and affine covariance', `temporal causality', `time-recursive updating mechanism' and `mutual consistency of an internal representation at different spatial and temporal scales'. He demonstrated that these types of receptive field models that cover spatial, spatio-chromatic and spatio-temporal receptive field properties have a very close similarity to the cell recording in biological vision. These cover a range of ON-center OFF-surround and vice versa RFs in the fovea and the LGN , then simple cells with spatial directional preference in V1, and even space-time separable spatio-temporal receptive fields in the LGN and V1 within the same unified theory. Lindeberg \cite{Lindeberg13} demonstrates how these mathematical models of the linear receptive fields (derived from the Gaussian kernels and its derivatives and closely related operators) can model the nonlinearity behaviour in their responses.
\subsection{The Proposed Model}
\label{sec:2.1}
The features of our bioplausible approach are intended to model the characteristics of a human's early visual processing. Based on numerous physiological studies e.g. \cite{Field07,Gauthier09,Gollisch10} there is a diversity of receptive field types and sizes inside the retina, resulting in multiscale encoding of the visual scene. This retinal representation is believed to be ``scale invariant'' in general, and there is an adaptation mechanism for the Receptive Field sizes to some textural elements (pattern features) inside our field of view \cite{Romeny08,Craft07}.

Marr and Hildreth \cite{Marr80} are well known for proposing the idea of multiscale edge analysis for producing edge maps of the image at different scales. In their proposal they argued that the optimal smoothing filter for images should be localized in both spatial and frequency domains. Gaussian filter is the only operator that satisfies this uncertainty principle \cite{Basu02} and has the best trade-off between the localization in both spatial and frequency domains simultaneously. The Gaussian operator in two dimensions is given by the following formula:
\begin{equation}
G(x,y)={\frac{1}{2\pi {\sigma}^2}}e^{-{\left(x^2+y^2\right)}/{\left(2\sigma^2\right)}} 
\label{equ:Gaussian}
\end{equation}
\noindent where $\sigma$ is the standard deviation of the Gaussian function (referred to as scale) and $(x,y)$ are the Cartesian coordinates of the image. They have suggested that by applying Gaussian filters at different scales to an image, a set of images with different levels of smoothness can be obtained \cite{Marr80,Basu02}, these are referred to as edge maps. In order to detect the edges in the edge map the zero-crossings of their second derivatives need to be found. Marr and Hildreth achieved this by applying the \emph{Laplacian of a Gaussian} function described  below:
\\[0.2mm]
\begin{eqnarray}
\nabla^2 G(x,y)={\frac{d^2}{dx^2}}G(x,y)+{\frac{d^2}{dy^2}}G(x,y)=\nonumber \\
{\frac{\left(x^2+y^2-2\sigma^2\right)}{2\pi {\sigma}^6}} e^{-{\left(x^2+y^2\right)}/{\left(2\sigma^2\right)}} 
\label{equ:Laplacian}
\end{eqnarray}
\\[0.2mm]
\noindent where $G(x,y)$ is the Gaussian function with standard deviation of $\sigma$, and $(x,y)$ are the Cartesian coordinates of the image. The LoG can be estimated as the Differences of two DoGs, estimating the second derivative of Gaussian as given in Eq. \eqref{equ:Laplacian}. For the DoG filtering, Cope et al., \cite{Cope13} used the following equation to represent the difference of two circular Gaussians:
\begin{equation}
f(x,y)={\frac{1}{2\pi {\sigma_c}^2}}e^{-{\left(r^2\right)}/{\left(2\sigma_c^2\right)}} - 
{\frac{\beta_{cs}}{2\pi {\sigma_s}^2}}e^{-{\left(r^2\right)}/{\left(2\sigma_s^2\right)}} 
\label{equ:DoG-Cope}
\end{equation}
\noindent Here $r$ is the radial distance from the center of the receptive field and is equal to $\sqrt{x^2+y^2}$, $\sigma_c$ is the center sigma, $\sigma_s$ is the surround sigma, and the parameter $\beta_{cs}$ is noted to be the dimensionless `balance parameter' in their model that has a range between $0<\beta_{cs}\le 1$. They have explained that the total volume of $f$ is $1-\beta_{cs}$, and $f$ is said to be balanced if $\beta_{cs}=1$ (results in a zero total volume of the filter). 

In our model, we have used simple circular DoG filters with zero total volume in the sense of \cite{Cope13}. These are thus perfectly `balanced', and we refer to these filters as `normalized DoGs'. The condition $0<\sigma_c<\sigma_s$ ensures that the regions of excitation is positioned in the center and therefore surrounded by the inhibition region of the surround Gaussian.

For an input image $I(x,y)$ with non-negative values, the DoG filters actually map the input pattern $I$ to output pattern $O$, using a `convolution operator':
\begin{equation}
O(\sigma_c,\sigma_s,…)=\int_{R×R}{f_{DoG} (\sigma_c,\sigma_s,…)I(x,y)dxdy}
\label{equ:mapping}
\end{equation}

For a 2D signal such as image $I$, the DoG output modelling the retinal ganglion cell responses with center surround organization in our implementation is given by:
\begin{eqnarray}
\Gamma_{\mathrm{\sigma,s\sigma}}(x,y)=I\ast{\frac{1}{2\pi {\sigma}^2}}e^{-{\left(x^2+y^2\right)}/{\left(2\sigma^2\right)}} - \nonumber \\
I\ast{\frac{1}{2\pi {s\sigma}^2}}e^{-{\left(x^2+y^2\right)}/{\left(2s^2\sigma^2\right)}} 
\label{equ:DoG}
\end{eqnarray}
\\[0.2mm]
\noindent where $x$ and $y$ are the distance from the origin in the horizontal and vertical axes respectively and $\sigma$ is the standard deviation/scale of the center Gaussian ($\sigma_c$). This is exactly the same as Eq. \eqref{equ:DoG-Cope} considering $\beta_{cs}=1$, for a normalized/zero total volume filter, and the size ratio of the center and surround Gaussians. As shown in Eq. \eqref{equ:s-factor}, $\sigma_s$ indicates the standard deviation (or scale) of the surround Gaussian ($\sigma_s=s\times\sigma_c$). $s$ is referred to as \emph{Surround ratio} in our model.
\begin{equation}
s=\sigma_\mathrm{surround}/\sigma_\mathrm{center}=\sigma_\mathrm{s} / \sigma_\mathrm{c}
\label{equ:s-factor}
\end{equation}

As indicated in Eq. \eqref{equ:DoG} by increasing the value of s, we reach a wider area of surround suppression, although the height of the surround Gaussian declines. We also tested a broader range of \emph{Surround ratio} from 1.4 to 8 but this made little difference to our results. For modelling the receptive field of retinal ganglion cells, DoG filtering \cite{Enroth-Cugell66,Rodieck65} is a good approximation of LoG, if the ratio of dispersion of center to surround is close to 1.6 \cite{Earle93,Marr80} ($s=1.6\approx\varphi$, the ubiquitous `Golden Ratio'). 

In addition to the $s$ factor, the filter size is another parameter to be considered in the model. The DoG is only applied within a window in which the value of both Gaussians are insignificant outside the window (less than 5\% for the surround Gaussian). A parameter is defined called \emph{Window ratio} ($h$) to control window size. The size is determined based on this parameter ($h$) and the scale of the center Gaussian ($\sigma_c$) as given in Eq. \eqref{equ:h-factor}:
\begin{equation}
windowSize=h\times\sigma_{\mathrm{c}}+1
\label{equ:h-factor}
\end{equation}
\noindent Parameter $h$ determines how much of each Gaussian (center and surround) is included inside the DoG filter (+1 as given in Eq. \eqref{equ:h-factor} guarantees a symmetric filter). Variations of DoGs have been tested with different values of $h$ (2 $\le h \le$ 12) while keeping $\sigma_c$ and $s$ constant. For the experimental results, in order to capture both excitation and inhibition effects, the \emph{Window ratio} ($h$) is set to 8 in this paper.

\begin{figure*}
	\centering
	\includegraphics[width=0.85\textwidth]{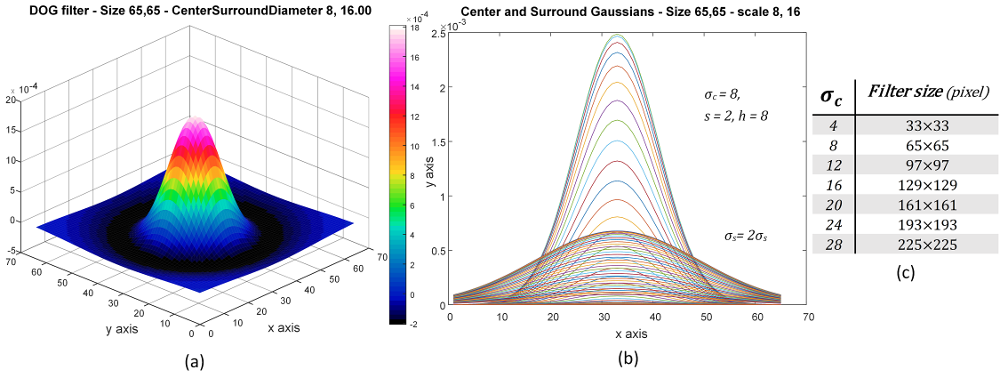}
	\caption{(a) 3D surface representation of Difference of Gaussian filters with standard deviations as $\sigma_c=8$ and $\sigma_s=16$ (\emph{Surround ratio} is 2.0), displayed in jetwhite colormap \cite{Powers16}. (b) 2D representation of the center and surround Gaussians for scale $\sigma_c=8$ and non-critical parameters of the model at $s=2$ and $h=8$ (\emph{Surround} and \emph{Window ratios} respectively). (c) The relationship between the scale of a DoG filter and the filter size, based on Eq. \eqref{equ:h-factor}.}
	\label{fig:2}   
\end{figure*}

A sample DoG filter is shown in \hyperref[fig:2]{Fig. \ref*{fig:2}}(a) for scale $\sigma_c=8$ and for the values of $s=2$, $h=8$. For more clarity, the center and surround Gaussians have been displayed in separate 2D-graphs in part (b) of the figure. \hyperref[fig:2]{Fig. \ref*{fig:2}}(c) shows the relationship between the scale of the center Gaussian in our model and the filter sizes at each scale that are defined based on Eq. \eqref{equ:h-factor} and the characteristics of the pattern. 

\begin{figure*}
  \centering
  \includegraphics[width=0.85\textwidth]{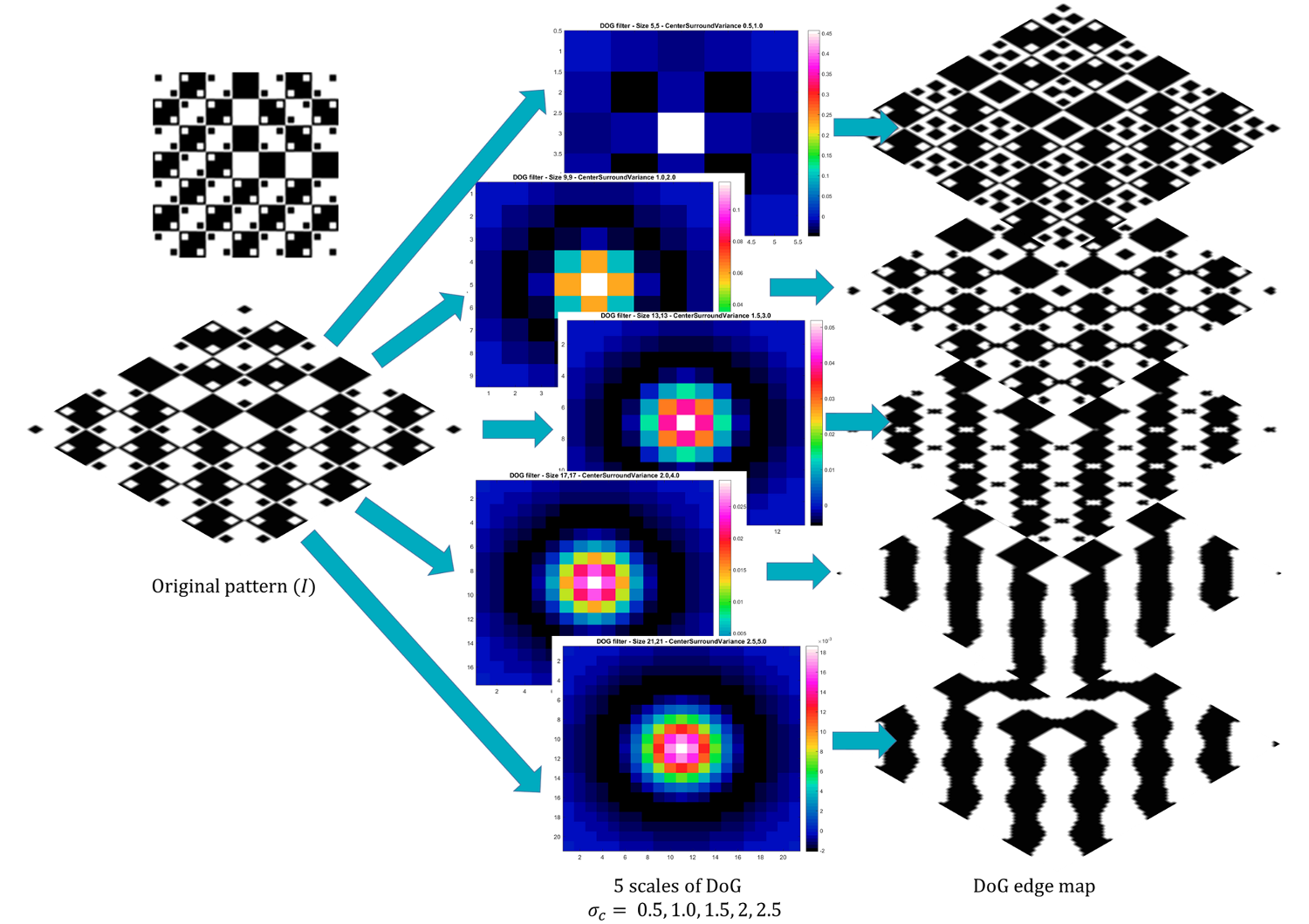}
  \caption{DoG edge map of a cropped section of the Trampoline pattern \cite{Kitaoka00} as a sample Tile Illusion with the tile sizes of 84$\times$84px. The scales of DoG filters are $\sigma_c=0.5$, 1.0, 1.5, 2.0, 2.5 to detect the important features in the pattern. Other parameters of the model are: $s=2$, and $h=8$ (\emph{Surround} and \emph{Window ratios} respectively).}
  \label{fig:3}   
\end{figure*}

A sample DoG edge map of a Tile Illusion is shown in \hyperref[fig:3]{Fig. \ref*{fig:3}}. The DoG is highly sensitive to spots of matching size, but is also responsive to lines of appropriate thickness and to contrast edges. A cropped section of the Trampoline pattern \cite{Kitaoka00} was selected as an input image (84$\times$84px). The edge map is shown at five different scales, $\sigma_c=0.5$, 1.0, 1.5, 2.0 and 2.5 to capture important features from the image (this is related to the texture/object sizes in the pattern and by applying Eq. \eqref{equ:h-factor}; we can determine a proper range for $\sigma_c$ in the model for any arbitrary pattern). The DoG filters in the figure are shown in the \emph{jetwhite colormap} \cite{Powers16}.

Scale invariant processing in general is not sensitive to the exact parameter setting; ideally the model's parameters should be set in a way that, at fine scales, they capture high frequency texture details and at coarse scales, the kernel has appropriate size relative to the objects within the scene.
The appearance of the Twisted Cord elements has previously been shown as the filtered output after either DoG or LoG on a Caf\'e Wall image at specific filter sizes (scales) \cite{Morgan86,Earle93}. We used a DoG derived model \cite{Nematzadeh15,Nematzadeh16a,Nematzadeh17a,Nematzadeh17b} that highlights the perception of divergence and convergence of mortar lines in the ``Caf\'e Wall'' illusion. We show that the model is capable of revealing the illusory cues in the pattern qualitatively, as well as allowing us to measure the strength and orientation of the tilt effect quantitatively in a wide range of Caf\'e Wall patterns.
\subsection{Edge map at multiple scales}
\label{sec:2.2}
The DoG transformation, modelling retinal ganglion cell (RGC) responses, creates an edge map representation at multiple scales for the pattern, consisting of a set of tilted line segments for the Caf\'e Wall stimulus \cite{Nematzadeh15,Nematzadeh16a,Nematzadeh17a,Nematzadeh17b} at its fine scales, noted as Twisted Cord elements in the literature \cite{Morgan86,Bressan85,Fraser08}. Then for quantitative measurement of tilt angles in the edge map to compare them with the tilt perceived by a human observer, we embed the DoG model in a processing pipeline using Hough space \cite{Nematzadeh16a,Nematzadeh16b}(details in \hyperref[appendix:AppxA]{Appendix A}).
 
The most fundamental parameter in the model is the scale of the center Gaussian ($\sigma_c$). Defining the scales in our model is highly correlated with characteristics of the pattern and its features, in particular the mortar lines and tile sizes in the Caf\'e Wall pattern. Based on the fixed parameters of the \emph{Surround and Window ratios}, relative to $\sigma_c$ ($s=2$ and $h=8$) and the pattern characteristics, an illustrative range for $\sigma_c$ to detect both mortar information and tiles is a range of $0.5\emph{M}=4$px to $3.5\emph{M}=28$px (refer to Eq. \eqref{equ:h-factor}). At scale 28 the DoG filter has a size of $8\times\sigma_c=8\times28=224$, close to Tile size$=200$px. Increasing the scale from this point results in a very distorted edge map, due to the DoG averaging and the filter size. We have selected incremental steps of 0.5\emph{M} between scales. This allows for the extraction of both mortar lines (at fine scales) and the Caf\'e Wall tiles (at coarse scales) in the edge map, as well as revealing the tilt cues and different perceptual groupings \cite{Nematzadeh17b} at multiple scales through the gradual increase of the DoG scales. So the output of the model is a DoG edge map at multiple scales for any arbitrary pattern. 

Since we have used normalized Gaussians in our model (\hyperref[sec:2.1]{Section \ref*{sec:2.1}}), the curve between the center and surround Gaussians in the DoG filter shows the activation response and that the surround Gaussian intersects the center Gaussian at its inflection points (for the 2.0 ratio). Therefore, $\sigma_c=4$ in our model corresponds to a filter in which the mortar size of 8px lies within one standard deviation ($\sigma_c$) away from the mean of the filter, so we are able to detect the mortar lines with high accuracy in the filtered response. Accordingly, the tilts detected near the horizontal at $\sigma_c=4$ in the quantitative tilt results (\hyperref[appendix:AppxC]{Appendix C}) show the \emph{prediction of tilt angles in the foveal vision} with high acuity in the center of our current gaze direction. We explain how to measure the slope of the detected tilted lines in the DoG edge maps in \hyperref[appendix:AppxA]{Appendix A}.

For a thorough investigation of tilt, we need to consider not just the horizontal mean tilts detected by our model from the edge maps at scale $\sigma_c=4$ that reflect the \emph{`foveal tilt effect'} (shown in red boxes in \hyperref[fig:C1]{Figs \ref*{fig:C1}}, \hyperref[fig:C2]{\ref*{fig:C2}}), but also more features from the edge maps and the predicted tilts by our model. 

One of the main scientific contributions of this study is to show a very important correlation that exists between the strength of the tilt effect in the Caf\'e Wall illusion and the persistence of mortar cues in the edge maps across multiple scales of the DoGs. We argue that the \emph{`persistence of mortar cues'} (PMC) plays a major role in determining how strongly we perceive the induced tilt in the Caf\'e Wall illusion. \hyperref[fig:5]{Fig. \ref*{fig:5}} shows the edge map for one of the stimulus tested ($ML=0.50$ in Mortar-Luminance variations). The PMC feature reflects the appearance of tilt and the generation of the Twisted Cord elements in the edge map of the Caf\'e Wall pattern that begins to emerge at scale 4 ($\sigma_c=4$) and persists till scale 16 ($\sigma_c=16$). In the next section, we will explain more about this feature, which we called PMC, and the information we can extract from the edge maps (\hyperref[sec:3.1.2]{Section \ref*{sec:3.1.2}}). 
\subsection{Patterns Investigated}
\label{sec:2.3}
To evaluate the tilts predicted by the model and to find out how close these predictions are to reported psychophysical experiments in the literature, we  generated different variations of the Caf\'e Wall pattern similar to the previously tested ones. All the generated patterns have the same configuration (number of rows and columns of tiles) as Caf\'e Walls of 3$\times$8 tiles with 200$\times$200px tiles as given in \hyperref[fig:4]{Fig. \ref*{fig:4}}.
The patterns investigated include \emph{Mortar-Luminance} (\emph{ML}) variations in which the mortar lines have a relative luminance in the range of Black ($\emph{ML}=0.00$) to White ($\emph{ML}=1.00$), and three shades of Grey in between Black and White (0.25, 0.50 and 0.75).

We also investigate \emph{Mortar-Width} (\emph{MW}) variations, ranging from no mortar lines ($\emph{MW}=0$px) through $\emph{MW}=4$, 8, 16, 32 and 64px.

The bottom of \hyperref[fig:4]{Fig. \ref*{fig:4}} show three patterns involving tiles with two shades of Grey (0.25 and 0.75) separated by mortar with one of three levels of luminance (0, 0.5, 1), and below that two variations showing different degrees of displacement (\emph{phase shifts} of 1/3 and 1/5), and finally a \emph{mirrored image} (\emph{Direction Change}) used to demonstrate that predicted tilts of the pattern reverses as expected and a \emph{hollow square} version that is used as an extreme case.
\section{Experimental Results}
\label{sec:3}
The tilt perception in the Caf\'e Wall pattern is not only affected by the foveal and peripheral view of the pattern \cite{Nematzadeh16a,Nematzadeh16b,Nematzadeh16c}, but also by the pattern characteristics such as mortar luminance, size, phase of tile displacement, tiles contrast and so forth. We analyse the effect of these parameters on the magnitude and orientation of the tilt effect in eighteen different variations of the Caf\'e Wall pattern (\hyperref[fig:4]{Fig. \ref*{fig:4}}).
\subsection{Mortar Luminance variations}
\label{sec:3.1}
\subsubsection{Pattern Characteristics}
\label{sec:3.1.1}
\begin{figure*}
  \centering
  \includegraphics[width=0.90\textwidth]{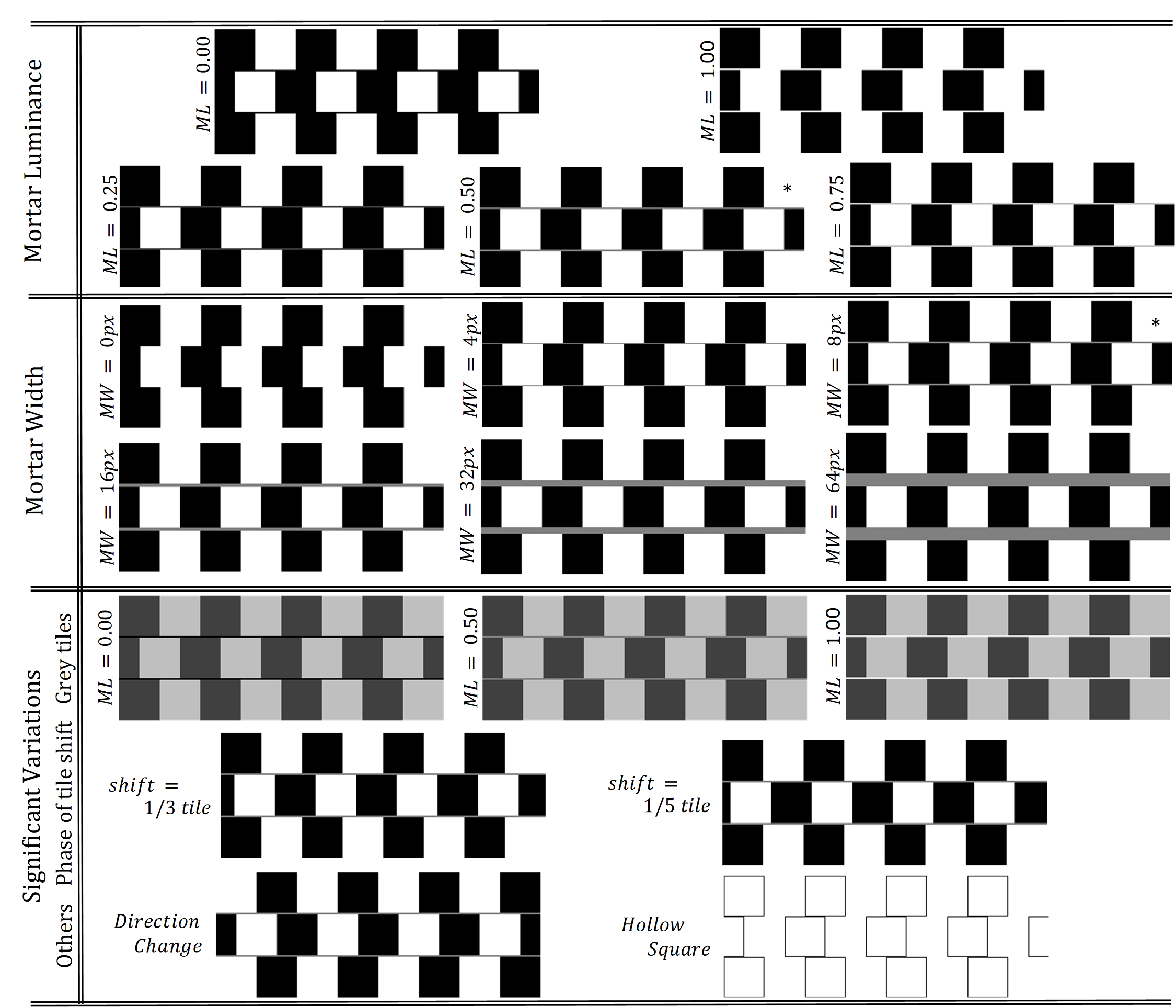}
  \caption{Patterns investigated. All the variations are Caf\'e Walls of 3$\times$8 tiles with 200$\times$200px tiles. Top: `Mortar Luminance' (\emph{ML}) variations from Black ($\emph{ML}=0.00$) to White ($\emph{ML}=1.00$) mortars, and three shades of Grey in between the Black and White ($\emph{ML}=0.25$, 0.50, and 0.75). Middle: `Mortar Width' (\emph{MW}) variations, from no mortar lines ($\emph{MW}=0$) and $\emph{MW}=4$, 8, 16, 32, 64px patterns (*: The original Caf\'e Wall with Black and White tiles, the mortar of intermediate luminance between the luminance of the tiles, and the $mortar width=8$px in our samples). Bottom: `Significant variations' investigated from Grey-Tiles with three luminance levels for mortar lines, below, between, or above the luminance of tiles, then phase of tile displacement for shifts of 1/3 and 1/5 of a tile between consecutive rows of the Caf\'e Wall, and finally mirrored image (Direction Change) inducing opposite direction of tilt as well as the Hollow Square pattern. Reproduced by permission from \cite{Nematzadeh18}.}
  \label{fig:4}   
\end{figure*}
\begin{figure*}
  \centering
  \includegraphics[width=0.7\textwidth]{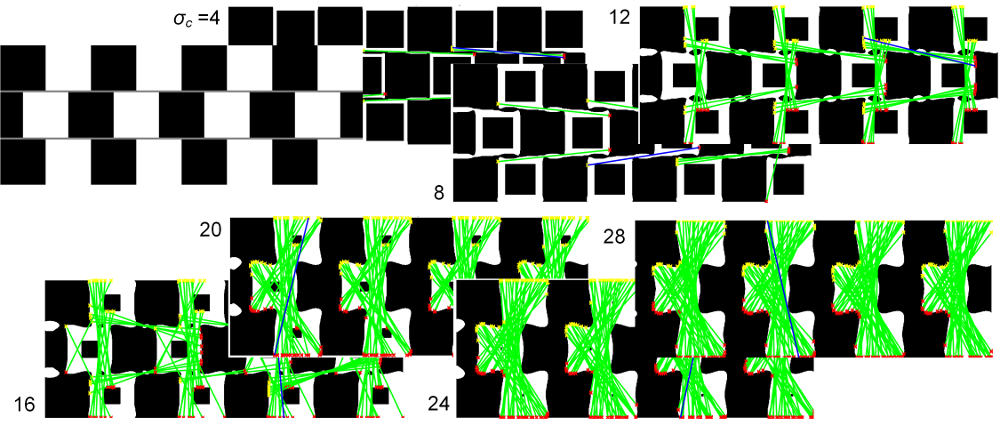}
  \caption {The binary DoG edge map at seven scales ($\sigma_c=4$, 8, 12, 16, 20, 24, 28) for the variation of $\emph{ML}=0.50$ (200$\times$200px tiles and 8px mortar). The edge map has been overlayed by \textsf{Hough lines} displayed in green (See \hyperref[appendix:AppxA]{Appendix A} for the procedure behind the extraction of the \textsf{Hough lines}).}
  \label{fig:5}  
\end{figure*}
The patterns under investigation are five variations given in \hyperref[fig:4]{Fig. \ref*{fig:4}}-top. These patterns are Caf\'e Walls with 200$\times$200px tiles (\emph{T}) and 8px mortar (\emph{M}).  To generate the samples, we have used a value that can be interpreted as `reflectance' (viewed printed) or `relative luminance (viewed on screen) to represent grey levels in the range [0,1]. Together with ambient direct and indirect illumination, these combine to give a subjective effect of relative brightness. Kingdom \cite{Kingdom11} defines `brightness' as the perceptual correlate of the perceived luminance. Blakeslee and McCourt \cite{Blakeslee15} explain that `luminance' is the physical intensity of a stimulus and that in achromatic vision, we see patterns of varying luminance, in which the judgement of `brightness', `lightness' and `contrast' would be identical in this case. In this paper we refer to the grey level of mortar lines as simply `luminance', and denote Mortar Luminance with \emph{ML} for an easier referral to this pattern characteristic in the rest of this paper.

By looking at the Mortar-Luminance variations, we see very minor tilt effects for the Black ($\emph{ML}=0.00$; the Munsterberg pattern) and White ($\emph{ML}=1.00$) mortar patterns, compared to the Grey mortar variations (with three luminance levels: $\emph{ML}=0.25$, 0.50, 0.75) and we investigate the predicted tilts by our model for these patterns.
\subsubsection{DoG Edge Maps and Quantitative Tilt Results}
\label{sec:3.1.2}
The binary DoG edge map at seven scales for $\emph{ML}=0.50$ in this category is given in \hyperref[fig:5]{Fig. \ref*{fig:5}}. The range of DoG scales starts from $0.5\emph{M}=4$px and continues to $3.5\emph{M}=28$px with incremental steps of $0.5\emph{M}$ ($\sigma_c=4$, 8, 12, 16, 20, 24, 28; \hyperref[sec:2.2]{Section \ref*{sec:2.2}}). The edge map has been overlayed by \textsf{Hough lines} displayed in green (check \hyperref[appendix:AppxB]{Appendix B}-\hyperref[fig:B1]{Fig. \ref*{fig:B1}} for a complete set of edge maps for these patterns with the overlayed \textsf{Hough lines}. Blue lines indicate the longest lines detected). Note that for the detection of near horizontal tilted line segments, as the perceived mortar lines in the Caf\'e Wall illusion, the DoG scale should be close to the mortar size, $\sigma_c\sim M=8$px \cite{Nematzadeh16a,Nematzadeh16b}. We suggest that $\sigma_c=4$ is appropriate for predicting the \emph{foveal tilt effect} and $\sigma_c=8$ for predicting the tilt at the edge of the image in the periphery of the retina. Comparing the edge maps at coarse scales ($\sigma_c=20$, 24, 28) shows very similar DoG outputs in most the variations tested. This is because, at the coarse scales, the scale of the DoG is large enough to capture the tile information. 

The difference between the DoG edge maps of the investigated patterns is mainly at fine to medium scales ($\sigma_c=4$, 8, 12, 16). As shown in \hyperref[fig:5]{Fig. \ref*{fig:5}} at scale 16, we see a transient state between detecting near horizontal tilted line segments connecting tiles with the Twisted Cord elements along the mortar lines, to zigzag vertical grouping of tiles at the coarse scales, in a completely opposite direction. At this scale, we still see some mortar cues left in the edge map of this stimulus, while in others, those mortar cues completely disappear. 

A key factor in making a predictive model for the illusory tilt in the Caf\'e Wall pattern is to consider essential visual cues that contribute to our perception of tilt in this illusion. We show that one of these informative cues for a correct prediction of tilt here is the \emph{persistence of mortar cues} (PMC) in the edge maps across multiple scales. For the prediction of the illusory tilt along the mortar lines, the main focus is on near horizontal tilts and generally at fine to medium scales of the edge maps. 

To \emph{predict the tilt effect} in the Caf\'e Wall illusion, we need to investigate both the `edge maps' and the `detected tilt angles' across multiple scales of the edge maps. As given in \hyperref[appendix:AppxC]{Appendix C}, we use the predicted mean tilts in the edge maps to extract the \emph{foveal tilt effects} (we called FTE) which is the detected tilt angles at scale 4 ($\sigma_c=4$, highlighted in red in \hyperref[fig:C1]{Figs \ref*{fig:C1}}, \hyperref[fig:C2]{\ref*{fig:C2}}), and also the \emph{range of detected tilt angles} at fine to medium scales (highlighted in blue boxes). Also from the DoG edge maps (\hyperref[appendix:AppxB]{Appendix B}), we specify the \emph{persistence of mortar cues} (PMC) across multiple scales. If no mortar cues are detected in the edge maps at scale 4, and similarly no \textsf{Hough lines} are detected in the Hough processing stage (\hyperref[appendix:AppxC]{Appendix C}), then we set the PMC feature to \emph{None} and the foveal tilt angle to zero ($FTE=0.0^{\circ}$). If the PMC lasts till  scale 8, the PMC is set to \emph{ML} (Medium-Low). If the PMC continues till the next scale ($\sigma_c=12$), we set the PMC value to \emph{M} (Medium), and for strong tilt effects with long-lasting PMCs reaching scale 16, we set the PMC to \emph{MH} (Medium-High). Above this scale ($\sigma_c>16$), if the PMC continues from medium to coarse scales, we set this feature to \emph{H} (High). 

To extract the PMC feature, an easy option is to use the mean tilts provided in \hyperref[fig:C1]{Figs \ref*{fig:C1}}, \hyperref[fig:C2]{\ref*{fig:C2}} (instead of checking the edge maps given in \hyperref[appendix:AppxB]{Appendix B}). For this, we should consider the range of scales from $\sigma_c=4$ till the scale with maximum mean tilt. For a scale to be considered with a maximum tilt angle, the minimum increment of tilt from the previous scale should be at least 1$^{\circ}$. We highlight the range of mean tilts in blue boxes in these figures that also indicate the range of corresponding scales for evaluating the PMCs in the Caf\'e Walls tested (for samples with no detected \textsf{Hough lines} at scale 4, the PMC is set to None and the FTE to $0.0^{\circ}$).
\begin{figure*}
  \centering
  \includegraphics[width=\textwidth]{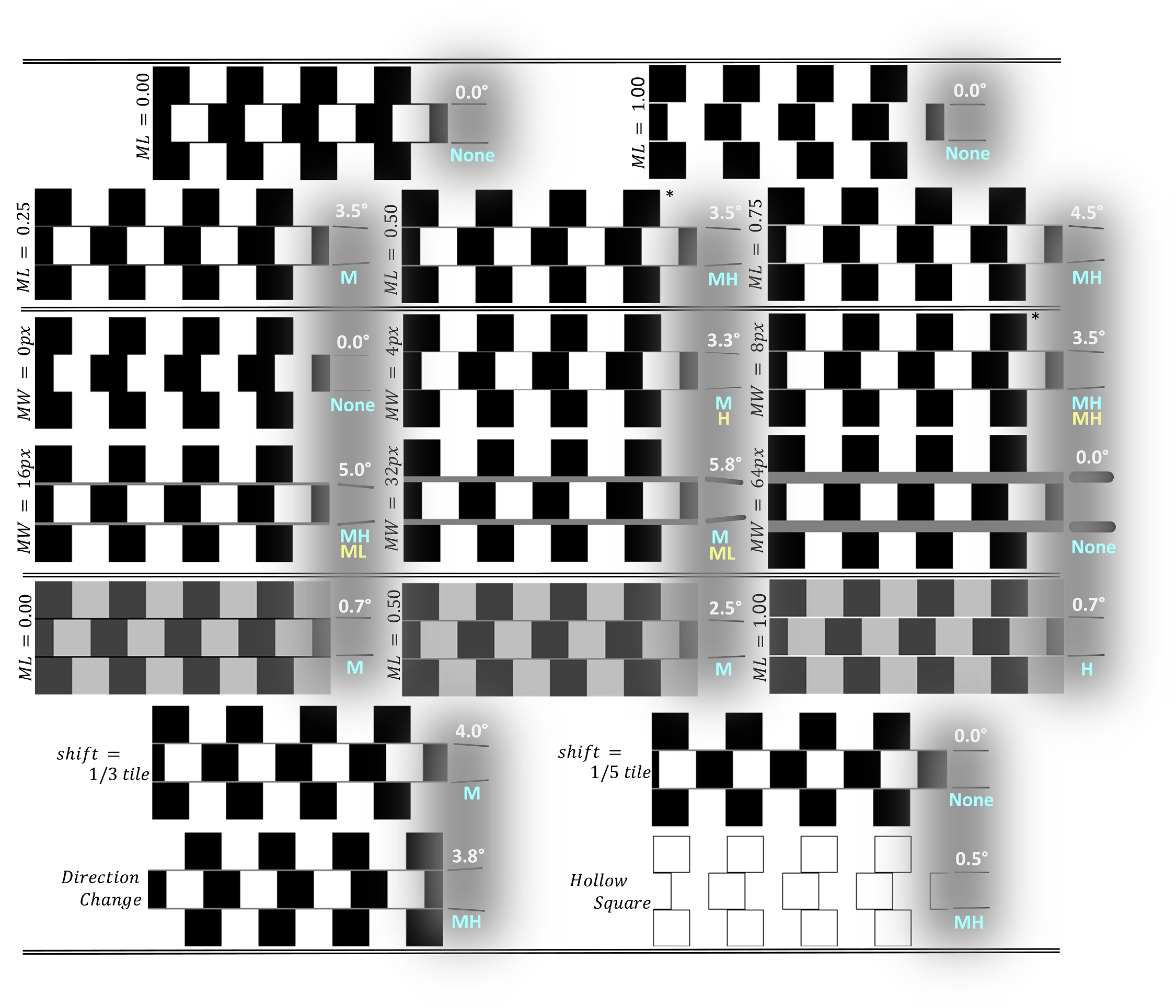}
  \caption{The summarized output of the model showing the `predicted tilt results' for the illusory tilt effect in the Caf\'e Wall patterns tested (Fig. 4). The results are provided in the Grey shaded areas next to each stimulus and on their right hand side. We have shown the \emph{`foveal tilt effect'} (FTE) in degrees (\emph{White}), which is the mean tilt angles predicted by our model at scale 4 of the edge maps. We have also presented a `qualitative feature' we called `\emph{persistence of mortar cues}' (PMC) that indicates the strength of the tilt effect in these patterns (\emph{Blue}), and their modified/corrected versions for the \emph{Mortar-Width} variations (\emph{Yellow}). The PMC feature can have values of \emph{None}, \emph{L}: Low, \emph{ML}: Medium-Low, \emph{M}: Medium, \emph{MH}: Medium-High and \emph{H}: High, based on how strongly the mortar cues are presented in the DoG edge maps and across multiple scales.}
  \label{fig:6}   
\end{figure*}

\hyperref[fig:6]{Fig. \ref*{fig:6}} shows the \emph{predicted tilt results} for these variations in Grey shaded areas, on the right hand side of each stimulus. This includes the foveal tilt effect (FTE) at scale 4 of the edge maps along the horizontal, presented in \emph{White} (from red boxes in \hyperref[fig:C1]{Figs \ref*{fig:C1}}, \hyperref[fig:C2]{\ref*{fig:C2}}), as well as the PMC values, presented in \emph{Blue} (from the corresponding range of scales shown by blue boxes in \hyperref[fig:C1]{Figs \ref*{fig:C1}}, \hyperref[fig:C2]{\ref*{fig:C2}}). We will explain later how the PMC values for the \emph{Mortar-Width} variations will be modified in relation to the mortar width, as shown in \emph{Yellow} in the figure. We have not reflected the \emph{range of tilt angles} at fine to medium scales in the figure and whenever it is required for our explanations, we refer to the Appendix (C). We have rounded up the mean tilts to one decimal point and eliminated the standard errors of calculations by considering only the horizontal deviations for this figure.
  
As the mean tilts in \hyperref[fig:C1]{Fig. \ref*{fig:C1}} show, there are no near horizontal lines detected at any of the DoG scales in the Munsterberg pattern ($\emph{ML}=0.00$). Based on the predicted tilt result shown in \hyperref[fig:6]{Fig. \ref*{fig:6}} (top-section) with a zero foveal tilt effect ($FTE=0.0^{\circ}$) and a \emph{None} PMC value, we note that there is no tilt illusion in the Munsterberg pattern ($\emph{ML}=0.00$).
For the White mortar variation, there are no detected lines at the finest scale ($\sigma_c=4$) similar to the Munsterberg pattern. But a few lines are detected around 1$^{\circ}$ of horizontal deviation at the next scale ($\sigma_c=8$) as per the details are given in \hyperref[fig:C1]{Fig. \ref*{fig:C1}}. If there is an illusion in the White mortar pattern, the tilts predicted by the model are quite negligible. Based on the predicted tilt results in \hyperref[fig:6]{Fig. \ref*{fig:6}} with $FTE=0.0^{\circ}$ and $PMC=None$, we conclude that as with the Black mortar, there is no illusion in the White mortar variation as well.

For the Dark-Grey ($\emph{ML}=0.25$) and Mid-Grey ($\emph{ML}=0.50$) mortars, the foveal tilt effects (FTE) are $\sim$3.5$^{\circ}$ compared to $\sim$4.5$^{\circ}$ for the $\emph{ML}=0.75$ (Light-Grey), as shown in \hyperref[fig:6]{Fig. \ref*{fig:6}}. The PMC values show that we are still able to detect horizontal lines at scale $\sigma_c=16$ for two patterns of $\emph{ML}=0.50$ and $\emph{ML}=0.75$ ($PMC=MH$), but not for $\emph{ML}=0.25$ ($PMC=M$). Here we need to consider the range of detected tilt angles at fine to medium scales (\hyperref[fig:C1]{Fig. \ref*{fig:C1}}) for a reliable tilt prediction in our investigations. We note that the range of detected mean tilts is up to 1$^{\circ}$ more for the $\emph{ML}=0.50$ compared to the $\emph{ML}=0.75$ pattern (between $\sim$3.5$^{\circ}$ to 12$^{\circ}$ compared to $\sim$4.5$^{\circ}$ to 12$^{\circ}$ at fine to medium scales). This could be an indication for a stronger tilt effect in the Mid-Grey pattern compared to the Light-Grey variation. This supports previous psychophysical findings that the highest strength for the tilt effect in the Caf\'e Wall illusion is when the luminance of the mortar is in the intermediate luminance of the tiles \cite{McCourt83,Morgan86}. We will explain later about the vertical and diagonal tilt angles detected by the model (as given in \hyperref[fig:C1]{Figs \ref*{fig:C1}}, \hyperref[fig:C2]{\ref*{fig:C2}}) for the phase displacement patterns, and here we keep the focus of our explanations on the tilt effects in the horizontal direction.
\subsection{Mortar Width variations}
\label{sec:3.2}
\subsubsection{Pattern Characteristics}
\label{sec:3.2.1}
The variations of Mortar-Width are shown in \hyperref[fig:4]{Fig. \ref*{fig:4}} and are denoted with \emph{MW} for easier referral to this pattern characteristic. We investigate following samples from the \emph{Mortar-Width} category: $\emph{MW}=0$, 4, 8, 16, 32, 64px. It has been reported that by increasing the mortar width, an inverse of the Caf\'e Wall illusion happens \cite{Earle93}, which is discussed here.
\subsubsection{DoG Edge Maps and Quantitative Tilt Results}
\label{sec:3.2.2}
\begin{figure*}
  \centering
  \includegraphics[width=0.7\textwidth]{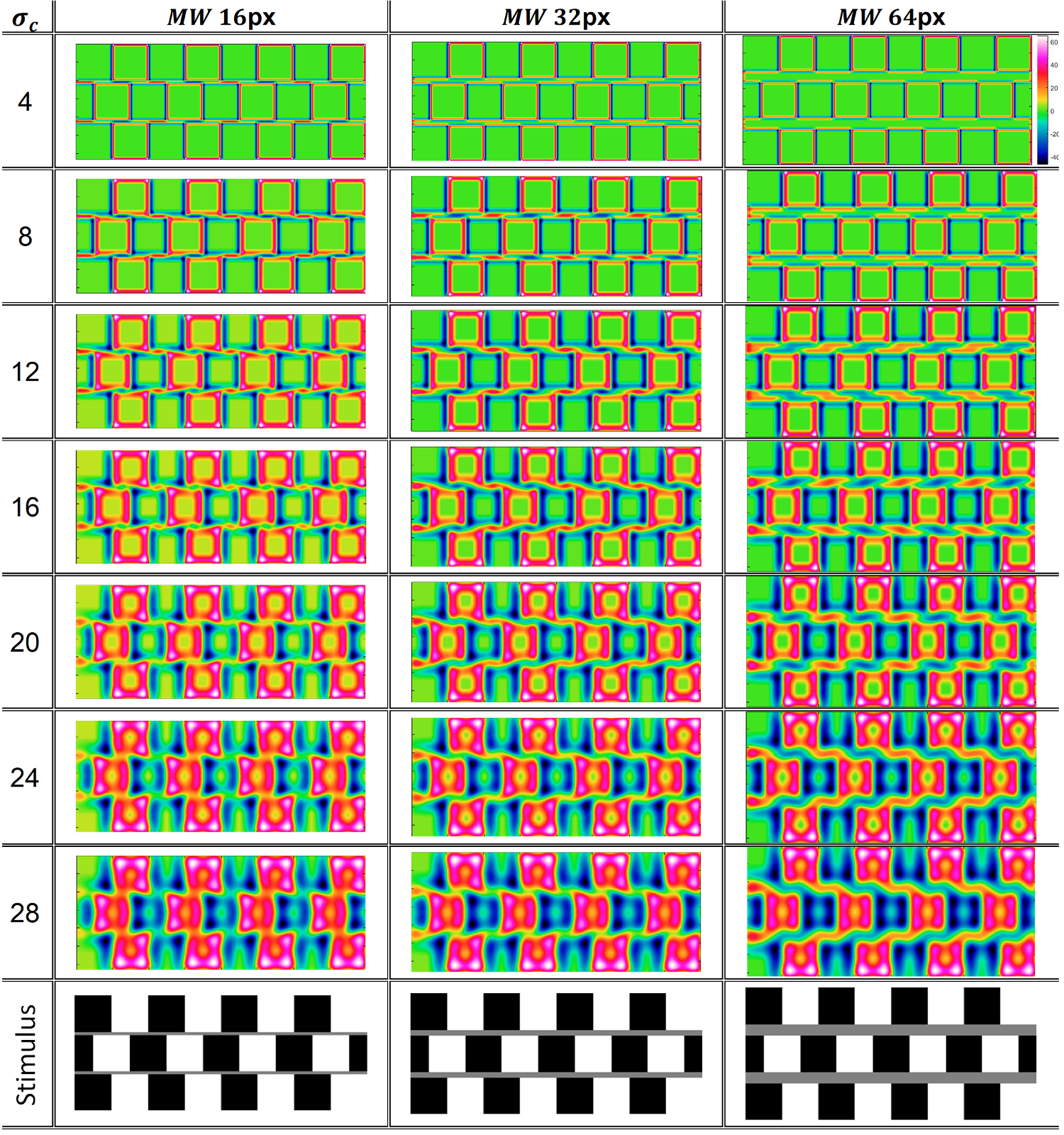}
  \caption{ DoG edge maps at seven scales ($\sigma_c$) for the thick mortar variations of $\emph{MW}=16$, 32, and 64px, displayed in the jetwhite colormap \cite{Powers16}. This figure highlights the connection between the `tilt effect' and the 'brightness induction' in the Caf\'e Wall illusions. This also shows how the Twisted Cord elements splits into two line segments and the underlying tilt effect and its direction in the thick mortar variations. The other DoG parameters of the model are $s=2$ and $h=8$ for the \emph{Surround} and \emph{Window ratios} respectively. Reproduced by permission from \cite{Nematzadeh18}.}
  \label{fig:7}   
\end{figure*}

Since the mortar widths are different in these variations while the tile sizes are the same, we have selected a pattern with mid-size mortar width, $\emph{MW}=8$px as a base, to define the scales of the DoG edge maps for these variations. So considering the mortar size of 8px (the DoG scales vary from 0.5\emph{M} to 3.5\emph{M} with incremental steps of 0.5\emph{M}), the appropriate scales are $\sigma_c=4$, 8, 12, 16, 20, 24, 28 which we use for all the Mortar-Width variations. \hyperref[fig:7]{Fig. \ref*{fig:7}} shows the DoG edge maps at seven scales in the jetwhite colormap for the $\emph{MW}=16$, 32 and 64px patterns (thick mortar lines). For a complete set of the DoG edge maps for these variations at fine to medium scales please refer to Appendix B (\hyperref[fig:B2]{Figs \ref*{fig:B2}}, \hyperref[fig:B3]{\ref*{fig:B3}}). 

As explained before, and due to the same sized tiles for all of the investigated patterns, the coarsest scales ($\sigma_c=24$, 28) show very similar DoG outputs. However, we see some substantial changes in the last two variations of very thick mortar lines ($\emph{MW}=32$, 64px) at these scales. The cues of mortar lines are still available for the $\emph{MW}=64$px pattern even at scale $\sigma_c=28$, but since no lines are detected at $\sigma_c=4$, the PMC value in \hyperref[fig:6]{Fig. \ref*{fig:6}} is set to \emph{None} for this pattern. The mortar in two patterns of $\emph{MW}=32$, 64px is quite thick, wide enough to separate the tiles completely, and our only perception when viewing these patterns, are the changes of brightness along the mortar lines. 
The Brightness Induction can be seen more clearly on the jetwhite colormap representations of the edge maps in \hyperref[fig:7]{Fig. \ref*{fig:7}} (also this can be seen in \hyperref[fig:B4]{Fig. \ref*{fig:B4}} in the binary edge maps). These brightness artifacts can be seen at scales 8 and 12 for the $\emph{MW}=32$px pattern and at scales 12, 16 and 20 for the $\emph{MW}=64$px. We want to emphasize that this is how \emph{`Brightness Induction'} connects to the \emph{`Tilt Effect'} in the Caf\'e Wall illusion and our explanation of illusory tilt. It becomes more apparent from \hyperref[fig:B3]{Figs \ref*{fig:B3}}, \hyperref[fig:7]{\ref*{fig:7}} that before the stages of Brightness Induction, we see breakage of the Twisted Cord elements into two parallel line segments at finer scales of the edge maps (at $\sigma_c=4$ for $\emph{MW}=32$px and at $\sigma_c=4$, 8 for $\emph{MW}=64$px). At these scales, the directions of groupings of identically coloured tiles are not as clear as they were in other patterns with thinner mortar lines (comparing \hyperref[fig:B2]{Figs \ref*{fig:B2}}, \hyperref[fig:B3]{\ref*{fig:B3}}). We refer to this effect as `Enlarged Twisted Cord'. The directions of these brightness changes are similar to what is detected at their fine DoG scales. The thick mortar variations have a strong brightness induction, but not a strong tilt effect.

In the literature it has been shown that when the diameter of the DoG operator is larger than the mortar width, an opposite phase brightness induction appears \cite{Foley85}. This has been reported as a Reverse of Caf\'e Wall illusion by Earle and Maskell \cite{Earle93}. Lulich and Stevens \cite{Lulich89} also reported: ``a reversal of the traditional Caf\'e Wall effect that is dependent upon mortar width'' (p. 428). The `Reversed Twister Cord' in thick mortar variations of the pattern is also called `Twisted Rope' \cite{Earle93} with an opposite direction to the Twisted Cord elements along mortar lines to distinguish these two. The term `Enlarged Twisted Cord' emphasizes the continuity with the Twisted Cord unlike the usage of Twisted Rope by Earle and Maskell \cite{Earle93}. They note that the effect of Reversed Twisted Cord occurs for a limited range of spatial frequency that is acting as bandpass filters. Outside this limit, the Twisted Cord elements break into two parallel segments aligned with the mortar lines \cite{Earle93}, and thus no Twisted Cord elements are presented. The breakage of Twisted Cords have been shown in the edge map representation explained above for $\emph{MW}=32$, 64px. The brightness induction observed and explained here is also consistent with Morgan and Moulden's conclusion \cite{Morgan86} that the effect is a consequence of Bandpass filtering. 

Lulich and Stevens \cite{Lulich89} note that by increasing mortar width the induced tilt along the mortar is further diminished and disappears when the mortar width is about twice the size of the DoG operator. In our results, the diminishing of the tilt cues can be viewed for the $\emph{MW}=8$px pattern at scale 16 of the edge map, but varies in other samples, with a higher range for the $\emph{MW}=4$px, and a much lower range for the thicker mortar lines (from the $\emph{MW}=16$px upwards). The scale of the DoG edge maps in the Mortar-Width variations does not directly relate to the mortar width of each individual stimulus in this class. This certainly affects our evaluations for the PMC feature for these patterns.

We should note that for the very thick mortar patterns ($\emph{MW}=32$, 64px), if the PMC feature is extracted from the DoG edge maps (\hyperref[fig:B3]{Figs \ref*{fig:B3}}, \hyperref[fig:B4]{\ref*{fig:B4}}), we get a different value compared to using the mean tilt tables (\hyperref[fig:C1]{Fig. \ref*{fig:C1}}). Therefore for this we use the detected mean tilts and the specified rule we defined in \hyperref[sec:3.1.2]{Section \ref*{sec:3.1.2}}. As explained before, we should consider a scale with maximum tilt angle considering that the minimum increment of tilt from the previous scale is at least 1.0$^{\circ}$. These PMC values are shown in Blue in \hyperref[fig:6]{Fig. \ref*{fig:6}} and we will explain later the necessity for modifying the PMC values in the Mortar-Width variations. 

As shown in \hyperref[fig:C1]{Fig. \ref*{fig:C1}}, there are no detected \textsf{Hough lines} around the horizontal orientation at any scales for $\emph{MW}=0$px. Here, the only grouping of pattern features that can be seen in the edge map across multiple scales are the zigzag vertical groupings of tiles (\hyperref[fig:B2]{Figs \ref*{fig:B2}}, \hyperref[fig:C1]{\ref*{fig:C1}}). Thus based on the predicted tilt results shown in \hyperref[fig:6]{Fig. \ref*{fig:6}} (middle-section) with a 0.0$^{\circ}$ FET and a \emph{None} PMC, we conclude that there is no tilt illusion in this pattern.

The foveal tilt effects (FTE) for the Mortar-Width variations as shown in \hyperref[fig:6]{Fig. \ref*{fig:6}} indicate that by increasing mortar width from 4px to 32px, there will be an increase of the detected tilt angles from approximately 3.3$^{\circ}$ in the $\emph{MW}=4$px pattern to roughly 5.8$^{\circ}$ in the $\emph{MW}=32$px variation. After this point we reach a 0.0$^{\circ}$ FTE and a \emph{None} PMC value for the $\emph{MW}=64$px. The reason for detecting larger tilt angles in the edge maps of thick mortar variations is that due to positioning thick mortars in between the shifted rows of tiles, the space between the same coloured tiles with the inducing tilt effect from the Twisted Cord elements increases as the the mortar width increases. This results in larger tilt angles that are correctly predicted by the model. This can be seen clearly in \hyperref[fig:B3]{Figs \ref*{fig:B3}}, \hyperref[fig:B4]{\ref*{fig:B4}} and \hyperref[fig:7]{\ref*{fig:7}}. This supports previous findings that in Caf\'e Wall patterns with thick mortar, the tilt inducing bands of Twisted Cords appear at a steeper angle to the horizontal compared to thinner mortar \cite{Lulich89}.

For the thickest mortar variation ($\emph{MW}=64$px) with the PMC value of \emph{None} and a 0.0$^{\circ}$ foveal tilt, we can conclude that there is no tilt effect in the pattern. The range of mean tilts of the thick mortars are larger than the range of horizontal mean tilts in the thinner mortar variations at fine to medium scales of the DoGs (which is $\sim$3.5$^{\circ}$- 9$^{\circ}$ in the $\emph{MW}=4$px, $\sim$3.5$^{\circ}$- 12$^{\circ}$ in the $\emph{MW}=8$px and $\sim$5$^{\circ}$- 14$^{\circ}$ in the $\emph{MW}=16$ and 32px patterns). When we compare the tilt effect in $\emph{MW}=64$px pattern with the thinner mortar variations such as the $\emph{MW}=8$px, we see that the tilt effect is very weak here and we see dominant brightness induction on the mortars. Checking the edge maps of the patterns with the overlayed \textsf{Hough lines} in \hyperref[fig:B3]{Fig. \ref*{fig:B3}} reveals that at scales 12 and 16, we have both negative and positive tilt angles detected along each mortar line. This also can be seen in the edge map of $\emph{MW}=32$px at $\sigma_c=8$. In the real vision, these contradictory tilts tend to cancel each other and result in a lower tilt range than the predicted tilt results.

As shown in \hyperref[fig:6]{Fig. \ref*{fig:6}}, the PMC values for the thick mortar variations for $\emph{MW}=16$, 32px are \emph{M} and \emph{MH} respectively, and the foveal tilt effects are the largest tilt angles detected in our samples. Despite these results, we do not see a strong tilt effect in them compared to the $\emph{MW}=4$, 8px. We will investigate further the very thick mortar variations in the next section to address this issue and to show how we are going to modify the PMC values for these variations to obtain a reliable tilt prediction. 
\subsubsection{Very Thick mortar variations}
\label{sec:3.2.3}
In the experiments reported so far, we have assumed the common hypothesis that for detection of near horizontal tilted line segments along mortar lines or the appearance of Twisted Cord elements in the literature, the DoG scale ($\sigma_c$) should be close to the mortar size ($\sigma_c$ $\sim$M ) \cite{Morgan86,Nematzadeh16a,Nematzadeh16b,Nematzadeh16c,Nematzadeh17a,Lulich89}. 
We show now that this is not precisely true when mortar size exceeds 16px in our samples (the Caf\'e Wall stimuli have tiles of 200$\times$200px). For two patterns of $\emph{MW}=32$ and 64px this hypothesis is not valid. We show here that the mortar cues have completely disappeared in the DoG edge maps of these patterns at scales much smaller than the mortar size (see \hyperref[fig:7]{Figs \ref*{fig:7}}, \hyperref[fig:B4]{\ref*{fig:B4}}).

For very thick mortar variations ($\emph{MW}=32$, 64px), we found that in the defined range of scales for the DoGs, the mortar cues still exist at scale 28. In addition, we have detected brightness induction in the DoG edge maps of these variations, much stronger than patterns with thinner mortar lines although the perception of tilt in these variations is very weak. To predict the tilts in these variations based on the summarized features in \hyperref[fig:6]{Fig. \ref*{fig:6}}, we see that despite our perception of weak tilt effect, the prediction solely upon the FTE shows strong tilts. The FTE actually shows the angles related to the appearance of the Twisted Cord elements in the edge maps and its steeper angles for the thick mortars  is being supported by previous reports \cite{Lulich89} (\hyperref[sec:3.2.2]{Section \ref*{sec:3.2.2}}). 

As indicated in \hyperref[fig:7]{Figs \ref*{fig:7}} and \hyperref[fig:B4]{\ref*{fig:B4}} for the $\emph{MW}=64$px pattern, we have mortar cues at scale 28 and a large deviation from the horizontal at this scale along the mortar lines. The zigzag vertical grouping of tiles which appeared clearly at coarse scales of the DoG edge maps for the $\emph{MW}=16$ and 32px patterns are not shown for the $\emph{MW}=64$px pattern in the predefined range of DoG scales. So we have examined a few scales above 28 for these patterns, and gathered some of the important results in \hyperref[fig:B4]{Fig. \ref*{fig:B4}} in an extended range of scales for the edge maps. We see a change of groupings of tiles from the near horizontal to the zigzag vertical at scale 24 for the $\emph{MW}=16$px pattern, and at scale 32 for the $\emph{MW}=32$px pattern, this happens around scale 48 for the $\emph{MW}=64$px (The emergence of the zigzag vertical groupings of tiles are highly correlated to the PMC feature).

The other thing worth mentioning from \hyperref[fig:B4]{Fig. \ref*{fig:B4}} is that the mortar lines are detectable at scale 16 ($\sigma_c=M$) for the $\emph{MW}=16$px pattern, but not at scale 32 ($\sigma_c=M$) for the $\emph{MW}=32$px pattern (they are detected at scale 24). There are no mortar cues available at scale $\sigma_c=64=M$ with the $\emph{MW}=64$px. In the edge map, the mortar cues exist till scale 32 ($\sigma_c=32$) and then disappear at coarse scales, with a filter size of half the mortar lines! This might be an indication for a very weak tilt effect, if there is any, for the $\emph{MW}=32$ and 64px patterns compared to the $\emph{MW}=16$px variation, despite the PMC values presented for them in \hyperref[fig:6]{Fig. \ref*{fig:6}} as \emph{M} and \emph{MH} respectively.

If we look back to the edge map of $\emph{MW}=8$px pattern (\hyperref[fig:B2]{Fig. \ref*{fig:B2}}), we see the mortar cues are not only detectable at scale 8 ($\sigma_c=M$), but also at the following two scales ($\sigma_c=12$, 16). The PMC value of \emph{MH} for this pattern (\hyperref[fig:6]{Fig. \ref*{fig:6}}) shows this. Comparing two variations of $\emph{MW}=8$px with $\emph{MW}=16$px (\hyperref[fig:B2]{Figs \ref*{fig:B2}}, \hyperref[fig:B3]{\ref*{fig:B3}}), we see that for the $\emph{MW}=16$px pattern, the mortar cues disappear in just one scale after the mortar size (at $\sigma_c=20$) except for very small dots in their place. Therefore, similar to $\emph{MW}=8$px, the PMC value for the $\emph{MW}=16$px is also \emph{MH}, but why the perceived tilt effect is weaker? For the very thin mortar ($\emph{MW}=4$px-\hyperref[fig:B2]{Fig. \ref*{fig:B2}}), the edge map shows persistent mortar cues from the finest scale ($\sigma_c=4=M$) till scale 12 and the PMC was set to \emph{M} in \hyperref[fig:6]{Fig. \ref*{fig:6}} (in Blue) although we see a very strong tilt effect in this pattern. Therefore, to keep this feature as a reliable indicator of the strength of the tilt effect, we need to modify the PMC values for the Mortar-Width variations. 

Using the same set of DoG scales for all the Mortar-Width variations to utilize the same set of Hough parameters for detecting tilt angles, resulting in unreliable PMC values for these stimuli. If we consider a proper range for the PMC feature in relation to the width of the mortar, then this problem will be fixed. For $\emph{MW}=4$px, the mortar cues last till scale 12 and this is actually three times of the mortar width in this pattern. Therefore, the PMC feature should be considered \emph{H} (Yellow - modified) instead of \emph{M} (Blue). The reason is that the PMC for the $\emph{MW}=4$px is even stronger than the PMC for the $\emph{MW}=8$px, that only last till scale $\sigma_c=16$ which is twice of the mortar size in the $\emph{MW}=8$px pattern. We have modified the PMC values in \hyperref[fig:6]{Fig. \ref*{fig:6}} with \emph{Yellow} colour considering the relation of the PMC to the mortar size for the Mortar-Width variations (and by comparing of the PMC feature of these variations with the original Caf\'e Wall pattern). By checking the modified PMC values (Yellow) for the Mortar-Width variations, we can see that the problem of \emph{overestimation} of the PMC values for the thick mortar lines ($\emph{MW}=16$, 32px), as well as the \emph{underestimation} of the PMC value for a very thin mortar lines ($\emph{MW}=4$px), are now resolved based on this strategy. 

Our DoG edge map representation supports the previous findings that the strongest tilt effect in the Caf\'e Wall illusion occurs with the thin mortar lines ($\emph{MW}=4$px with $PMC=H$ and then 8px with $PMC=MH$ in our samples). Also, the edge map at multiple scales reveals the underlying cues involved in thick mortar variations of the Caf\'e Wall illusion and indicates how the tilt effect degrades here while the brightness induction increases in these patterns. Our model seems to be a good fit for our biological understanding of the early stages of vision. 
\subsection{Significant Variations of the Pattern}
\label{sec:3.3}
One of the main factors in illusions is the brightness or lightness of the elements, and it is important to see whether the precise shade of squares is a significant factor. We have also checked removing the cell shades completely to verify that the illusion disappears. Similarly the illusions shown have all been based on 1/2 shift of tiles. This raises the question of how differently these shifts effect the illusion. These patterns are shown at the bottom of \hyperref[fig:4]{Fig. \ref*{fig:4}} which contains \emph{Grey Tiles} variations, \emph{Phase of tile displacement}, \emph{mirrored image} and \emph{Hollow Square}.
\subsubsection{Grey Tiles variations}
\label{sec:3.3.1}
Grey-Tiles are variations of Caf\'e Wall patterns with lower contrasted tiles, in which instead of the Black and White tiles with maximum contrast, the tiles here are two shades of Grey. The relative luminance of tiles in these variations are equal to 0.25 for Dark-Grey and 0.75 for Light-Grey). So in these variations the luminance contrast between the tiles is half of the luminance contrast of the original Caf\'e Wall pattern with the Black and White tiles. The mortar lines have one of the three levels of luminance here, either below, between, or above the luminance of both of the Grey tiles selected as $\emph{ML}=0.00$ (Black), 0.50 (Mid-Grey), and 1.00 (White). The foveal tilt effects (FTEs) and the PMC values for these variations are extracted from \hyperref[fig:C2]{Fig. \ref*{fig:C2}}, similar to the previously investigated patterns shown in \hyperref[fig:6]{Fig. \ref*{fig:6}}. 

For the Black mortar pattern ($\emph{ML}=0.00$), the foveal tilt effect is much lower than the original Caf\'e Wall, and close to the $\emph{ML}=1.00$. This is less than 1$^{\circ}$ for $\emph{ML}=0.00$ and $1.00$ (Black and White mortar) and $\sim$2.5$^{\circ}$ for Mid-Grey mortar. Although for the Grey-Tiles, the FTEs are smaller than the majority of the other variations tested with high contrasted tiles of Black and White tiles, their PMC features are \emph{M} to \emph{H} (\emph{M} for $\emph{ML}=0.00$, 0.50 and \emph{H} for $\emph{ML}=1.00$). It has been reported that the strength of illusion in low contrasted tile variations of the pattern is less than the original Caf\'e Wall with high contrasted tiles \cite{Gregory79,Morgan86}. Here we need to check the range of detected tilt angles at fine to medium scales to address the overall tilt effect in these patterns. \hyperref[fig:C2]{Fig. \ref*{fig:C2}} shows that the range of detected tilt angles for $\emph{ML}=0.50$ is between 2.5$^{\circ}$ to 7.5$^{\circ}$ (compared to 3.5$^{\circ}$ to 12$^{\circ}$ in the original Caf\'e Wall), that is a good indication for a weak tilt effect. Therefore checking the range of mean tilt angles (at fine to medium scales) seems to be essential in some cases to lead us towards a reliable prediction for the tilt effect in the Caf\'e Wall illusion. Again for the $\emph{ML}=1.00$, we find a lower range for the predicted mean tilts at fine to medium scales, indicating a weak tilt effect compared to the original Caf\'e Wall pattern. 

Gregory and Heard \cite{Gregory79} noted that the effect depends critically on the luminance and that it disappears if the mortar line is either brighter than the light tiles, or dimmer than the dark ones, and our results support these psychophysical findings. Also it has been reported that Caf\'e Walls with lower contrasted tiles need thinner mortar lines to have the same degree of apparent convergence/divergence in the illusion percept \cite{Gregory79}. This has not yet been tested by our model (But for Mortar-Width variations with contrast luminance of one unit for the tiles we have illustrated that patterns with thinner mortar lines produce a wider range of tilt angles indicating stronger tilt effect-\hyperref[sec:3.2]{Section \ref*{sec:3.2}}).
\subsubsection{Phase displacement effect on detected tilt }
\label{sec:3.3.2}
Two more patterns are generated based on two different phase of tile displacements with more details as follows. The phase of tile displacement is the amount of tile shift between consecutive rows in the Caf\'e Wall pattern. One of these patterns has a phase shift of 1/3 (shift of 1/3$^{rd}$ of a tile) and the other one a phase shift of 1/5. Since there is no near horizontal line detected along the mortar at the finest scale for the shift of 1/5 of a tile, we set the PMC feature to \emph{None} and the FTE to 0.0$^{\circ}$ (refer to \hyperref[sec:3.1.2]{Sections \ref*{sec:3.1.2}}). For the phase shift of 1/3, the foveal tilt effect is 4.0$^{\circ}$ and the PMC feature is \emph{M}. Checking the mean tilts at fine to medium scales ($\sigma_c=4$, 8, 12-\hyperref[fig:C2]{Fig. \ref*{fig:C2}}) for the phase shift of 1/3 with the original Caf\'e Wall pattern shows that we can detect a lower range of mean tilts which is $\sim$4$^{\circ}$-9.5$^{\circ}$ compared to $\sim$3.5$^{\circ}$-12$^{\circ}$, that highlights again a weaker tilt effect in this pattern compared to the original Caf\'e Wall. 

Now, we explain the effect of \emph{`vertical and diagonal mean tilts'} in these variations as an example in between our samples in this section and leave similar investigations for the rest of the patterns to the reader. As shown in \hyperref[fig:C2]{Fig. \ref*{fig:C2}}, the vertical mean tilts is much smaller in the pattern with the phase shift of 1/5 around 2.5$^{\circ}$-3.5$^{\circ}$ at different scales. It is slightly higher in the phase shift of 1/3 ($\sim$4$^{\circ}$-6$^{\circ}$); recall that for the near vertical and diagonal tilts, we should consider coarse scale DoGs in our model; $\sigma_c>16$. Checking the edge maps in \hyperref[fig:B6]{Fig. \ref*{fig:B6}} shows that the vertical grouping of tiles appears at scale 8 in the two variations of tile shift displacement compared to scale 12 and even 16 in the original Caf\'e Wall pattern. This is a good indication of a weaker tilt effect along the mortar lines in these variations compared to the original Caf\'e Wall because the tilt cues are more persistent along mortar lines at fine to medium scales in the original Caf\'e Wall pattern ($PMC=\emph{MH}$ in \hyperref[fig:6]{Fig. \ref*{fig:6}}). In addition, we see sharper vertical lines with less deviation from the vertical in the phase shift of 1/5, highlighting the vertical grouping of tiles that emerges at smaller scales. The near diagonal mean tilts for these two patterns (phase shifts of 1/3 and 1/5) are more than 3$^{\circ}$ lower than the predicted tilt ranges along the diagonal axes compared to the original Caf\'e Wall ($\sim$9.5$^{\circ}$ compared to $\sim$12.5$^{\circ}$-13.5$^{\circ}$-\hyperref[fig:C2]{Fig. \ref*{fig:C2}}). Our results are consistent with previous reports that the tilt effect is maximal with a phase shift of half a cycle between consecutive rows and minimal or no distortion when it is in a checkerboard phase \cite{McCourt83}.
\begin{figure*}
  \centering
  \includegraphics[width=\textwidth]{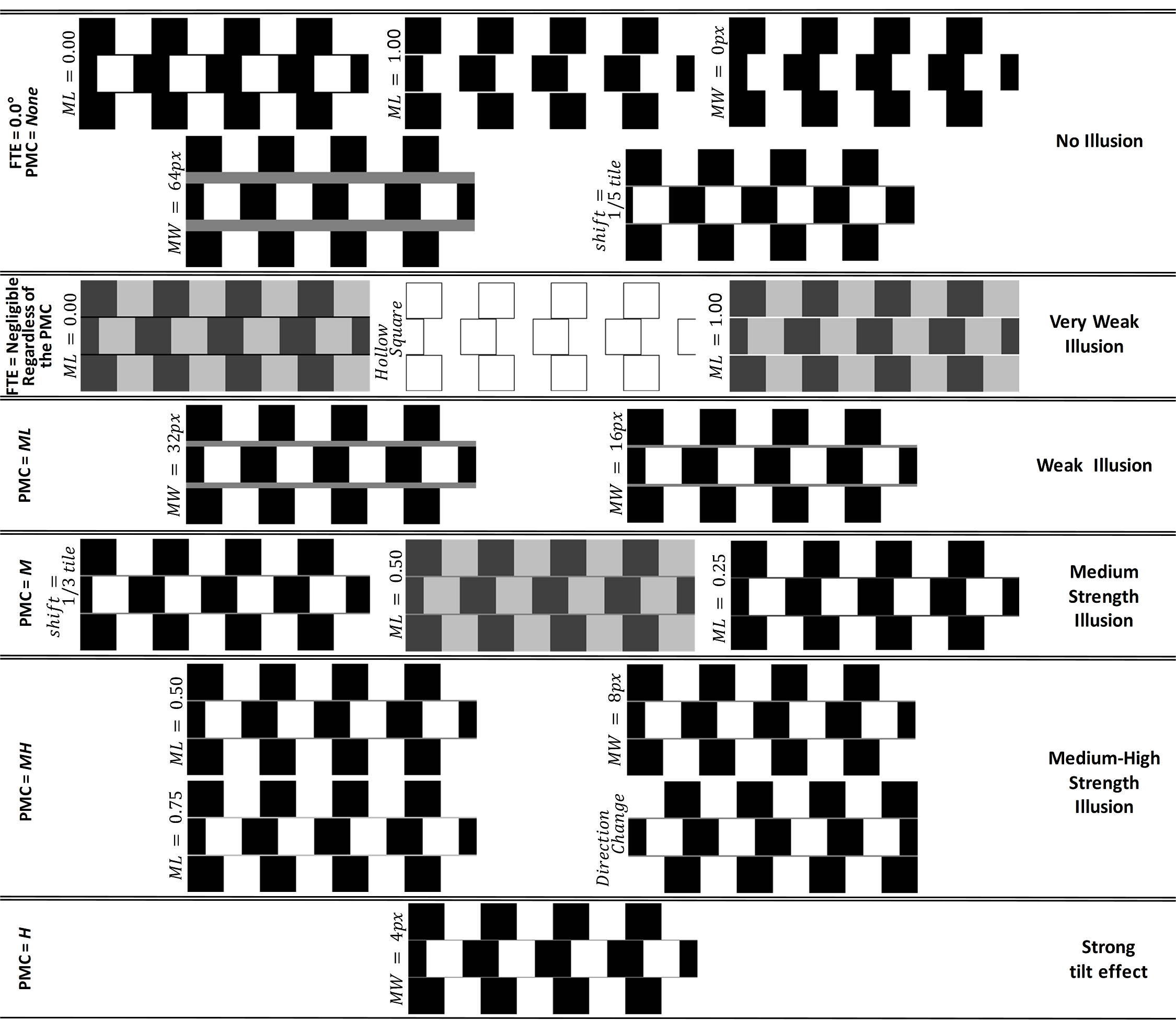}
  \caption{The predicted tilt effects for the Caf\'e Walls tested (Fig. 4) that are classified in an ascending order from `No illusion' to the `Strong tilt effect' based on the model predictions. The features on the left hand side of the figure that were used for classification are the \emph{foveal tilt effect} (FTE) in degrees  and the \emph{persistence of mortar cues} (PMC) in the range of {\emph{None, ML, M, MH, H}} as explained in Fig. 6.}
  \label{fig:8}   
\end{figure*} 
\subsubsection{Tilt effect direction and Hollow Square pattern}
\label{sec:3.3.3}
The last two patterns we investigate in this work are the \emph{mirrored image} and the variation of the \emph{Hollow Square} that are shown at the bottom of \hyperref[fig:4]{Fig. \ref*{fig:4}}. The \emph{mirrored image} of the original Caf\'e Wall pattern (which is the mirrored image of either $\emph{ML}=0.50$ or $\emph{MW}=8$px, asterisked in figure) results in an opposite direction of induced tilt. This version is referred to as the \emph{Direction Change} variation of the Caf\'e Wall in the figures. The final pattern we consider is the \emph{Hollow Square} pattern \cite{Woodhouse87,Bressan85}, which consists of hollow tiles with the same size of the Caf\'e Wall tiles. The outlines of tiles are thinner than the mortar size but the border thickness of these hollow tiles produces a roughly similar size to the mortar size of 8px in the Caf\'e Wall pattern, since two hollow tiles are adjacent to each other. If the outlines of the hollow squares are thickened in this version, we ultimately reach a similar pattern to the Caf\'e Wall without any mortar lines \cite{Woodhouse87}. As shown in \hyperref[fig:6]{Fig. \ref*{fig:6}} (bottom-section) for the Direction Change pattern (Mirrored Image), the foveal tilt effect (FTE) is quite similar to the original Caf\'e Wall pattern. Even the near horizontal tilt range is quite similar to the original Caf\'e Wall pattern as shown in \hyperref[fig:C2]{Fig. \ref*{fig:C2}}, ranging from $\sim$3.5$^{\circ}$ to 12$^{\circ}$ at fine to medium scales in the edge maps (shown in \hyperref[fig:6]{Fig. \ref*{fig:6}} with the PMC value of \emph{MH}). The mean tilts along the vertical and diagonal orientations are again very similar to the original Caf\'e Wall pattern. Slight changes less than a degree ($˂1^{\circ}$) are in the acceptable mean tilt range due to the standard errors around 0.5$^{\circ}$-0.6$^{\circ}$ in the original Caf\'e Wall, and this error indicates that the results are statistically very close to each other in these two variations. This is what we expected, and we have shown that the tilt effect has an opposite direction of divergence/convergence tilts in the detected \textsf{Hough lines} (\hyperref[fig:B7]{Fig. \ref*{fig:B7}}).

The last pattern we consider is the \emph{Hollow Square}. Comparing the foveal tilt effect for the Hollow Square (\hyperref[fig:6]{Fig. \ref*{fig:6}}) with the original Caf\'e Wall pattern shows that the FTE is negligible ($<1^{\circ}$) here compared to 3.5$^{\circ}$ in the original Caf\'e Wall pattern. Despite the weak tilt effect in this pattern, its PMC value is \emph{MH} that is overestimated. By further investigation of \hyperref[fig:B7]{Figs \ref*{fig:B7}}, \hyperref[fig:C2]{\ref*{fig:C2}}, we found that at scale 4, the mean tilt is quite negligible (0.5$^{\circ}\pm$0.5) and it is $\sim$6$^{\circ}$ at scale 8. But what is important to consider is that although the detected lines have a mean tilt deviation around 6$^{\circ}$, the lines have both positive and negative orientations along each mortar position at $\sigma_c=8$ (connecting of hollow tiles in rows) compared to a single tilt orientation (either positive or negative tilt) for the detected lines along each mortar in the original Caf\'e Wall pattern. In the real vision, at this transition resolution, these contradictory tilts tend to cancel each other and result in a lower tilt range than the predicted tilt results ($\sim$6$^{\circ}$).

We suggest that for the investigation of tilt in a more general class of Tile Illusions, we need to modify the measurement of tilt angles in the model to be able to consider the sign of detected angles in the calculations. Currently, for calculating the mean tilts in the model (Hough processing stage), we only consider the absolute values of the detected mean tilts. Among all the variations of the Caf\'e Walls tested, the Hollow Square and the very thick mortars of $\emph{MW}=32$ and 64px were the only patterns with both positive and negative tilt angles detected along the position of each mortar line (at scales 8 and 12). For the Hollow Square version, it has been also reported that the decrease of contrast reduces the apparent tilt of the \emph{Hollow Square} illusion \cite{Woodhouse87}. This has not yet been tested by our model.
\subsection{Summary for the predicted tilt effects}
\label{sec:3.4}
As shown in \hyperref[fig:8]{Fig. \ref*{fig:8}}, we have classified the tilts predicted by our model for all the variations tested based on the strength of the Caf\'e Wall illusion. We emphasize that all the features used for the investigation of tilt in the illusion (\hyperref[fig:6]{Fig. \ref*{fig:6}}), including the `foveal tilt effect' (FTE), the `persistence of mortar cues' (PMC) and the `range of detected mean tilts' at fine to medium scales (\hyperref[appendix:AppxC]{Appendix C}), are detected by the low- to mid-level model employed in our study. A systematic approach was designed as a classifier for the tilt effect that is described in the following:
\begin{itemize}
\item[o] If \underline{$FTE=0.0^{\circ}$} and $PMC=None$ $\Rightarrow$ `No illusion' (e.g. $ML=0.00$, 1.00 in Mortar-Luminance variations and $shift=1/5$tile)
\item[o] If \underline{$FTE$ is negligible} ($<1^{\circ}$), regardless of the $PMC$ value $\Rightarrow$ `Very Weak' tilt effect (e.g. $ML=0.00$ and 1.00 in the Grey-Tiles patterns)
\item[o] If \underline{$PMC=ML$} (for $FTE>1^{\circ}$) $\Rightarrow$  `Medium-Low' tilt effect (e.g. the Thick-mortar variations: $MW=16$px, 32px)
\item[o] If \underline{$PMC=M$} (for $FTE>1^{\circ}$) $\Rightarrow$ `Medium' strength tilt effect (e.g. $shift=1/3$tile and $ML=0.25$)
\item[o] If \underline{$PMC=MH$} (for $FTE>1^{\circ}$) $\Rightarrow$ `Medium-High' strength tilt effect (e.g. the original Caf\'e Wall: $ML=0.50$/$MW=8$px and the $Direction Change$/mirrored image)
\item[o] If \underline{$PMC=H$} (for $FTE>1^{\circ}$) $\Rightarrow$ `Strong' tilt effect (here we predicted that the $MW=4$px shows the strongest tilt effect in between our samples)
\end{itemize}

Therefore, the most substantial factor to detect `None' to `Very Weak' tilt effect is when $FTE=0.0^{\circ}$ or quite negligible ($<1^{\circ}$, such as $ML=1.00$ in Grey-Tile variations). Otherwise, for the Caf\'e Walls with apparent tilt effects (from weak to strong), we rely on the $PMC$ feature, as a key indicator to classify them based on the strength of the illusion. As previously explained for the Caf\'e Walls investigated, $FTE$ and `range of detected mean tilts' at fine to medium scales provide additional information to let us compare the tilt effects in different stimuli. For instance, in the category of `Medium-High' strength tilt effects (\hyperref[fig:8]{Fig. \ref*{fig:8}}), the $ML=0.50$/$MW=8$px both indicate the original Caf\'e Wall in our samples and they have similar tilt effects to the `Direction Change' pattern based on the detected features (provided in \hyperref[fig:6]{Fig. \ref*{fig:6}}). Also, $ML=0.75$ has been classified in this category but the `range of detected mean tilts' for this pattern shows a weaker tilt effect compared to the original Caf\'e Wall pattern.

We have made quantitative predictions of tilt for a wide range of conditions, and specifically, the predictions of the strength of illusion as conditions vary. For the first time, we have early stage vision that makes quantifiable predictions about a family of illusions.
\section{Conclusion}
\label{sec:4}	
It is increasingly clear that information in the visual systems is processed at multiple levels of resolution, perhaps simultaneously, perhaps sequentially in some sense. We examined the capability of a bioplausible vision model, simulating retinal/cortical simple cells to address the illusory percept in variations of the Caf\'e Wall pattern. Exploring the tilt effect in the Caf\'e Wall illusion, we have shown that a simple DoG model of lateral inhibition in retinal/cortical simple cells leads to the emergence of tilt in the pattern. Our model generates an intermediate representation at multiple scales that we refer to as an edge map. For the recognition of a line at a particular angle of tilt, further processing by orientation selective cells in the retina and/or cortex is assumed \cite{Grossberg85,Moulden79} but we have exploited an image processing pipeline for quantitative measurement of tilt angle using Hough transform. In our study, the possible location of edges in the edge maps is determined by using the Hough technique. This procedure is closely related to template matching, generates data concerning curve detection, i.e. line detection in the given solution \cite{Ballard82}. The biological evidence for template matching has been proposed by Hubel and Wiesel \cite{Hubel79} that the mammalian visual system responds to edges based on a special low-level template matching edge detectors. A further way where the model should be developed is to replace the use of the \textsf{Hough line} detection with the proposed successive layers of biologically plausible neurons, similar to the approach of Von der Malsburg \cite{Malsburg73}.

We have shown in this paper that the DoG edge map not only shows the emergence of tilt in the Caf\'e Wall illusion but can also explain different degrees of tilt effect in variations of the Caf\'e Wall illusion based on their characteristics. The qualitative and quantitative tilt results of the Caf\'e Wall variations investigated support previous psychophysical and experimental findings \cite{Gregory79,McCourt83,Morgan86,Earle93,Woodhouse87,Grossberg85,Moulden79,Foley85,Lulich89} on these stimuli.

We have shown that explanatory models and hypotheses for the Caf\'e Wall illusion such as \emph{Irradiation}, \emph{Brightness Induction}, and \emph{Bandpass filtering} appear to share the central mechanism of lateral inhibition that ultimately underlies the tilt effect in this illusory pattern. We further expect that these retinal filter models will prove to play an important role in higher–level models of simulating depth and motion processing. This supports the use of Gaussian Filters and their differences or derivatives in Computer Vision. We have also shown empirically that this model has a high potential to reveal the underlying mechanism connecting low-level filtering approaches to mid- and high-level explanations such as `Anchoring theory' and `Perceptual grouping' \cite{Nematzadeh17b}.

Although we have covered many of the aspects involved in the illusory tilt perceived in variations of the Caf\'e Wall pattern by our model in this work (through relying on the available psychophysical reports in the literature), many things are left to be explored. These include psychophysical experiments as a priority in our future study to confirm the predictions implicit in our results, and they are expected to lead us to a more precise multiple scale filtering which is adaptable to the pattern characteristics.
\renewcommand{\theHsection}{A\arabic{section}}
\setcounter{secnumdepth}{0}
\counterwithin{figure}{section}
\newcommand{\hbAppendixPrefixA}{A}
\renewcommand{\thefigure}{\hbAppendixPrefixA\arabic{figure}}
\setcounter{figure}{0}

\section{Appendix}
\addappheadtotoc
\begin{subappendices}
\section*{\\A. Quantitative Measurement of Tilt Angles-Analysis with Hough}
\label{appendix:AppxA}
\begin{figure*}[htp]
  \centering
  \includegraphics[width=0.85\textwidth]{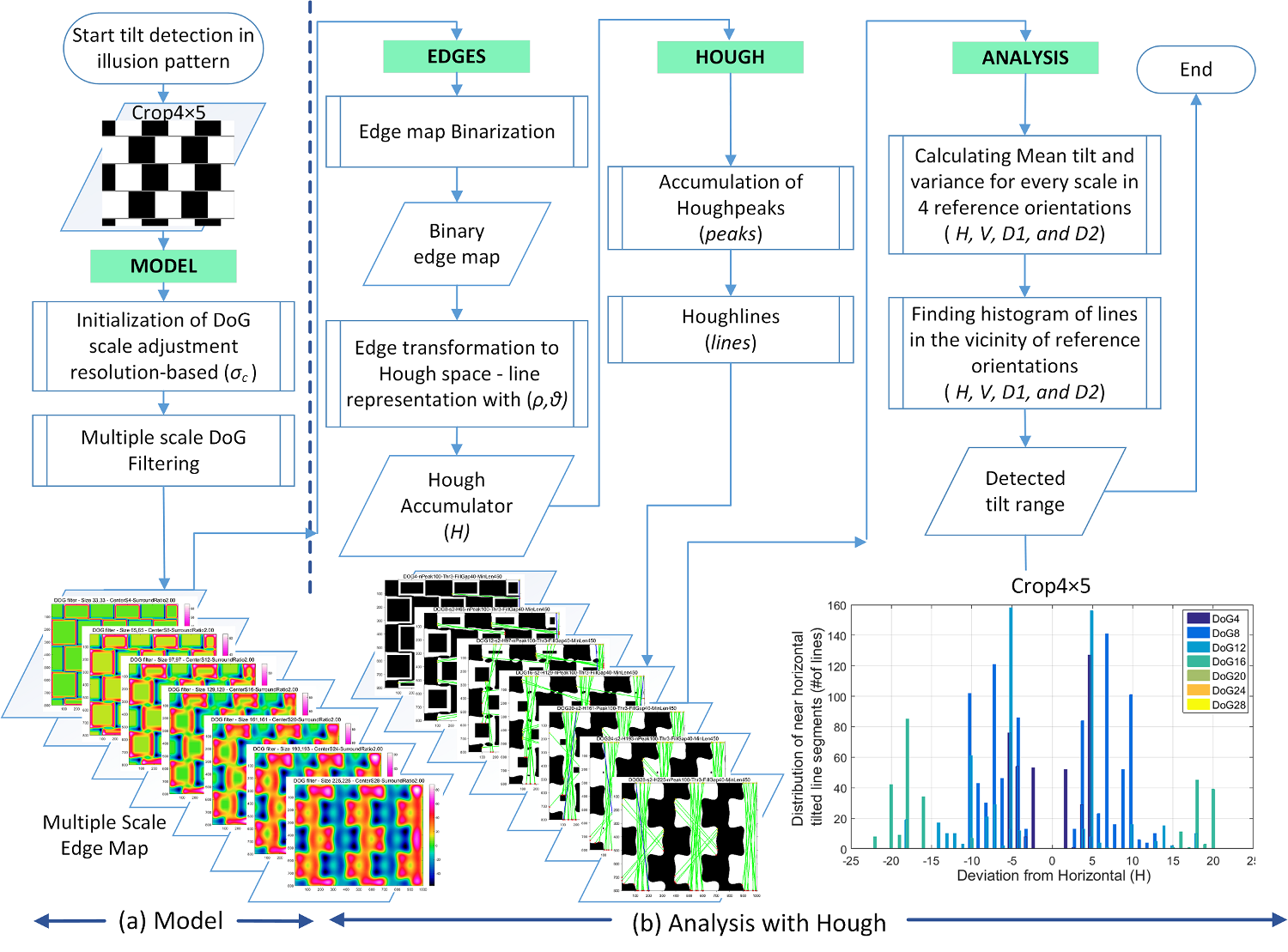}
  \caption{Flowchart of the model and analytical tilt processing. Reproduced with permission from [25].}
  \label{fig:A1}   
\end{figure*}
The DoG edge map at seven scales for a cropped section of a Caf\'e Wall pattern with 200$\times$200px Tiles (\emph{T}) and 8px Mortar (\emph{M}) is shown in \hyperref[fig:A1]{Fig. \ref*{fig:A1}}, as the output of the model, revealing the Twisted Cord elements along the mortar lines. The quantitative tilt measurement  includes three stages (\hyperref[fig:A1]{Fig. \ref*{fig:A1}}), \textsc{EDGES}, \textsc{HOUGH} and \textsc{ANALYSIS} implemented in MATLAB. 
\\[0.3cm]\noindent\textsc{EDGES}: At each scale, first the edge map is binarized and then Hough Transform (HT) [73] is applied which allows us to measure the tilt angles in detected line segments in the binary edge map. HT uses a two-dimensional array called the accumulator ($H_A$) to store lines information with quantized values of $\rho$ and $\theta$ in its cells. $\rho$ represents the distance between the line passing through the edge point, and $\theta$ is the counter-clockwise angle between the normal vector ($\rho$) and the x-axis, with the range of [0, $\pi$). So based on new parameters of $(\rho, \theta)$ every edge pixel $(x, y)$ in the image space corresponds to a sinusoidal curve in the Hough space as given by $\rho= x.cos\theta+y.sin\theta$.
\\[0.3cm]\noindent\textsc{HOUGH}: All possible lines that could pass through every edge point in the edge map, are accumulated inside the $H_A$ matrix. We are more interested in the detection of tilt inducing line segments inside the Caf\'e Wall pattern. Two MATLAB functions called \textsf{houghpeaks} and \textsf{houghlines} were employed for this reason. The \textsf{houghpeaks} function finds the peaks in the $H_A$ matrix, which are the dominant line segments. It has parameters of \textsl{NumPeaks} (maximum number of lines to be detected), \textsl{Threshold} (threshold value for searching the $H_A$ for the peaks), and \textsl{NHoodSize} (neighborhood suppression size which is set to zero after the peak is identified). The \textsf{houghlines} function, however, extracts line segments associated with a particular bin in the $H_A$. It has parameters of \textsl{FillGap} (maximum gap allowed between two line segments associated with the same Hough bin), and \textsl{MinLength} (minimum length for merged lines to be kept). Sample outputs of the \textsc{HOUGH} analysis stage are presented in \hyperref[fig:B1]{Figs \ref*{fig:B1}} to \hyperref[fig:B3]{\ref*{fig:B3}}, and \hyperref[fig:B5]{\ref*{fig:B5}} to \hyperref[fig:B7]{\ref*{fig:B7}} for different variations of the Caf\'e Wall pattern investigated (Detected \textsf{Hough lines} are shown in green, displayed on the binary edge maps).
\\[0.3cm]\noindent\textsc{ANALYSIS}: To categorize detected line segments, we have considered four reference orientations of horizontal (H), vertical (V), positive diagonal (+45$^{\circ}$, D1), and negative diagonal (-45$^{\circ}$, D2). An interval of [-22.5$^{\circ}$, 22.5$^{\circ}$) around each reference orientation was chosen to cover the whole space. The information from \textsc{HOUGH} is saved inside four orientation matrices based on how close they are to one of these reference orientations for further tilt analysis. The statistical tilt measurements of the detected \textsf{Hough lines} in the neighborhood of each reference orientation is the output of this final stage.

To quantify the mean tilts in the edge maps for these patterns using the Hough analysis pipeline in our model, the same set of parameters of Hough are applied for all of these variations, and for every scale of the edge maps to attain reliable tilt results, which are comparable between these variations. The fixed Hough parameters are: \textsl{NumPeaks}= 1000, \textsl{FillGap}= 40px, and \textsl{MinLength}= 450px.

\renewcommand{\theHsection}{B\arabic{section}}
\setcounter{secnumdepth}{0}
\counterwithin{figure}{section}
\newcommand{\hbAppendixPrefixB}{B}
\renewcommand{\thefigure}{\hbAppendixPrefixB\arabic{figure}}
\setcounter{figure}{0}

\section*{\\B. Detected Hough lines on the Edge Maps}
\label{appendix:AppxB}
\begin{figure*}[htp]
  \centering
  \includegraphics[width=\textwidth]{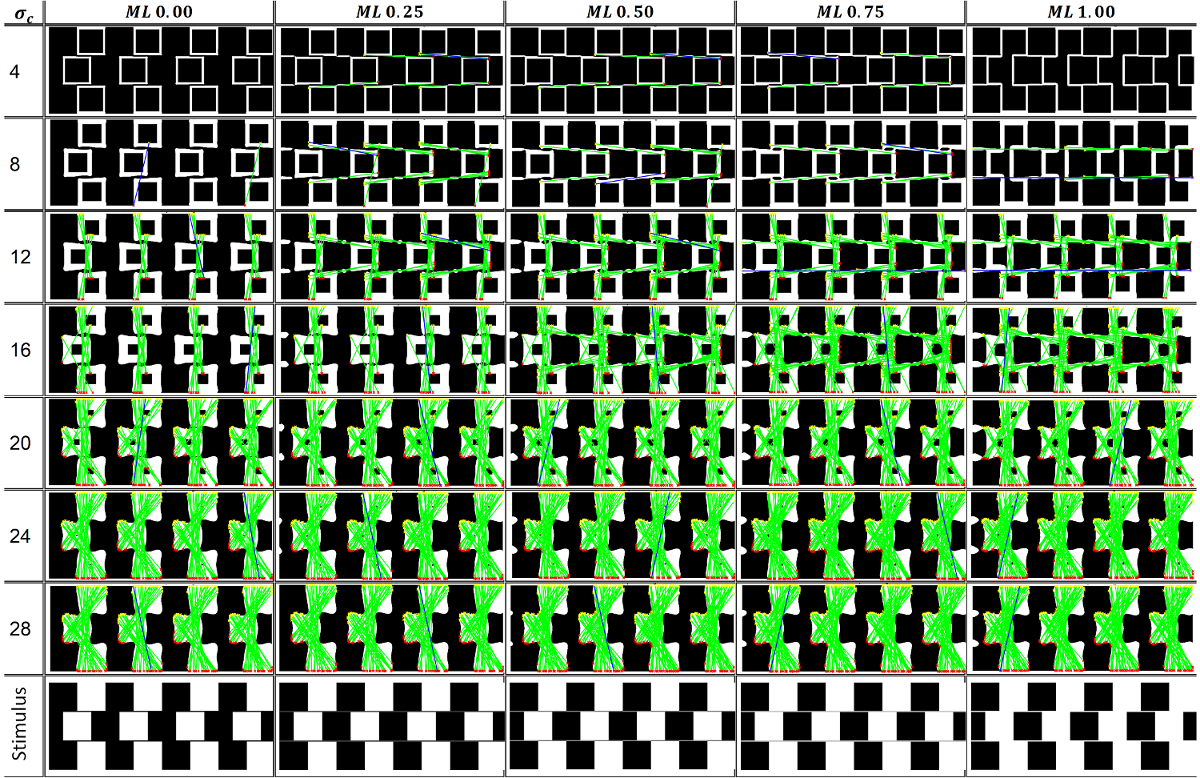}
  \caption{DoG edge maps at seven scales ($\sigma_c$) for five Mortar-Luminance variations, from Black (\emph{ML} 0.00) to White (\emph{ML} 1.00) mortar lines, in which the edge maps are overlayed by the detected \textsf{Hough lines} displayed in green (Blue lines indicate the longest lines detected). Hough parameters are kept the same in all experiments to detect near Horizontal, Vertical and Diagonal tilted lines in the edge maps (\hyperref[appendix:AppxA]{Appendix A}). Other parameters of the DoG model are kept constant at $s= 2$ (\emph{Surround ratio}), $h= 8$ (\emph{Window ratio}). The last row shows each variation of the pattern investigated. Reproduced by permission from [25].}
  \label{fig:B1}   
\end{figure*}
\begin{figure*}[htp]
  \centering
  \includegraphics[width=0.65\textwidth]{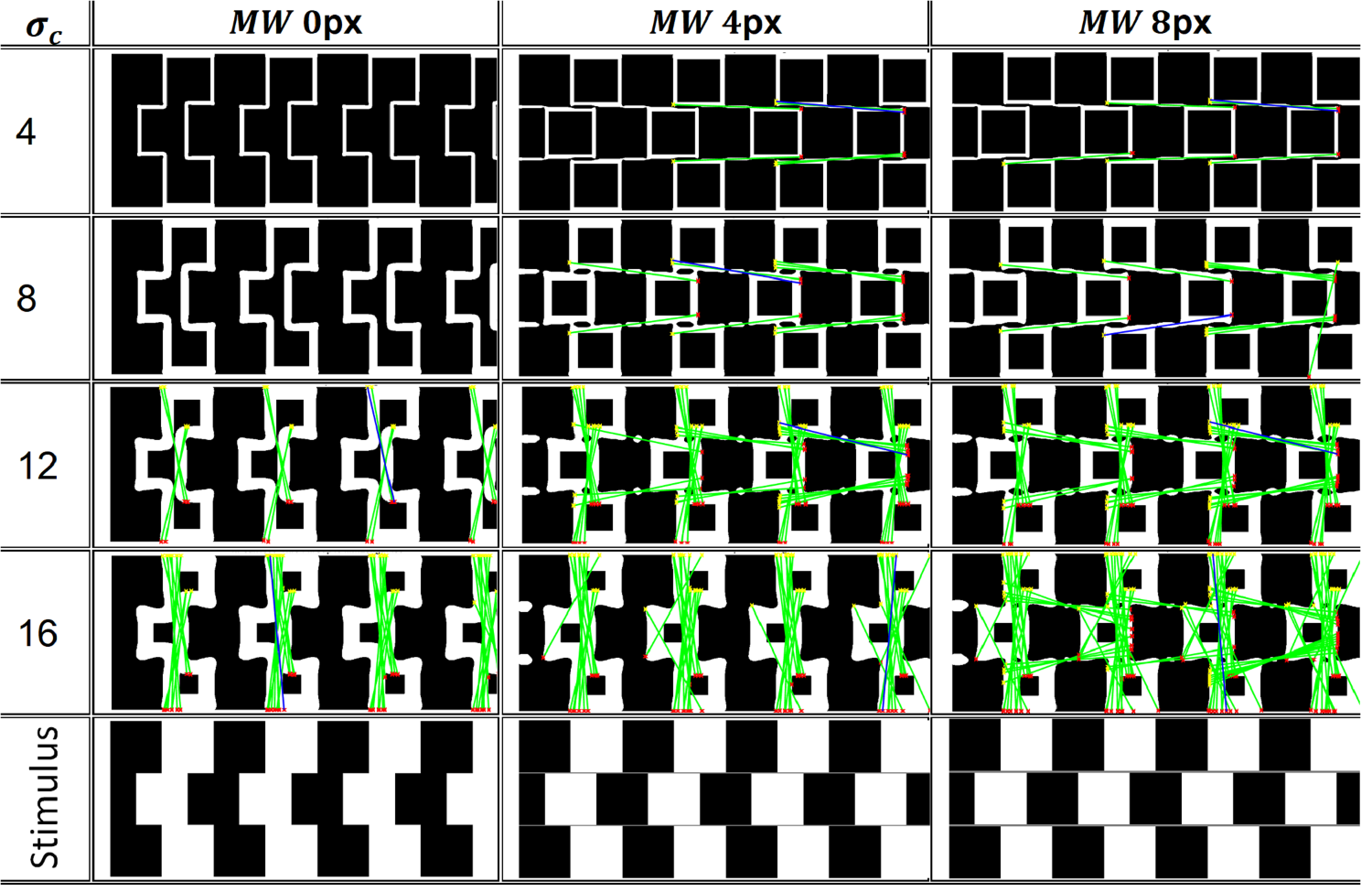}
  \caption{DoG edge maps at fine to medium scales ($\sigma_c$) for three variations of Mortar-Width, from no mortar lines (\emph{MW} 0px) to 8px mortar, with the overlayed \textsf{Hough lines} displayed in green. Other parameters of the model and Hough processing stage are the same as Fig. B1.}
  \label{fig:B2}   
\end{figure*}
\begin{figure*}[htp]
  \centering
  \includegraphics[width=0.65\textwidth]{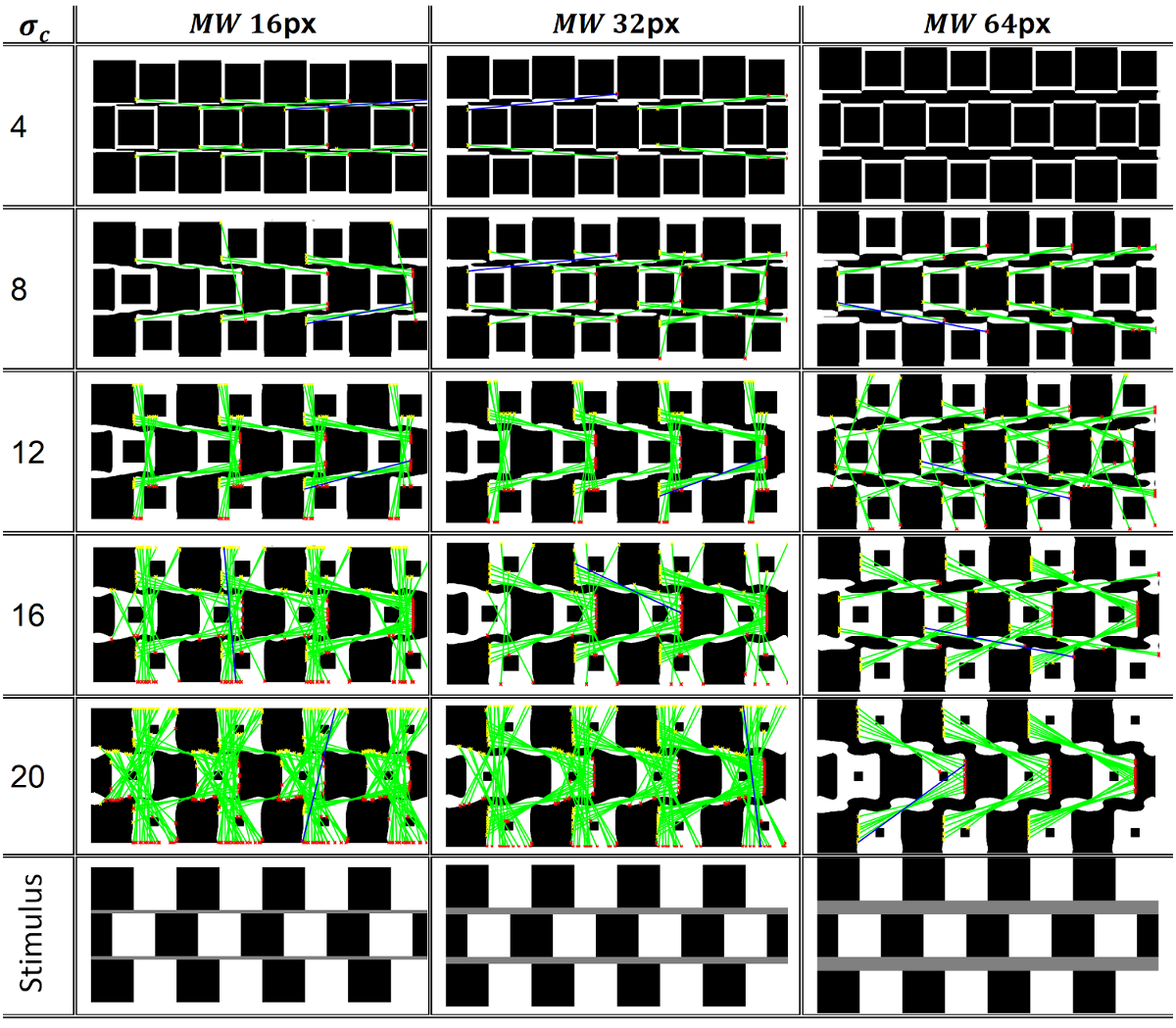}
  \caption{DoG edge maps at fine to medium scales ($\sigma_c$) for three variations of Mortar-Width, from mortar size of 16px to 64px, with the overlayed \textsf{Hough lines} displayed in green. Other parameters of the model and Hough processing stage are the same as Fig. B1.}
  \label{fig:B3}   
\end{figure*}
\begin{figure*}[htp]
  \centering
  \includegraphics[width=0.65\textwidth]{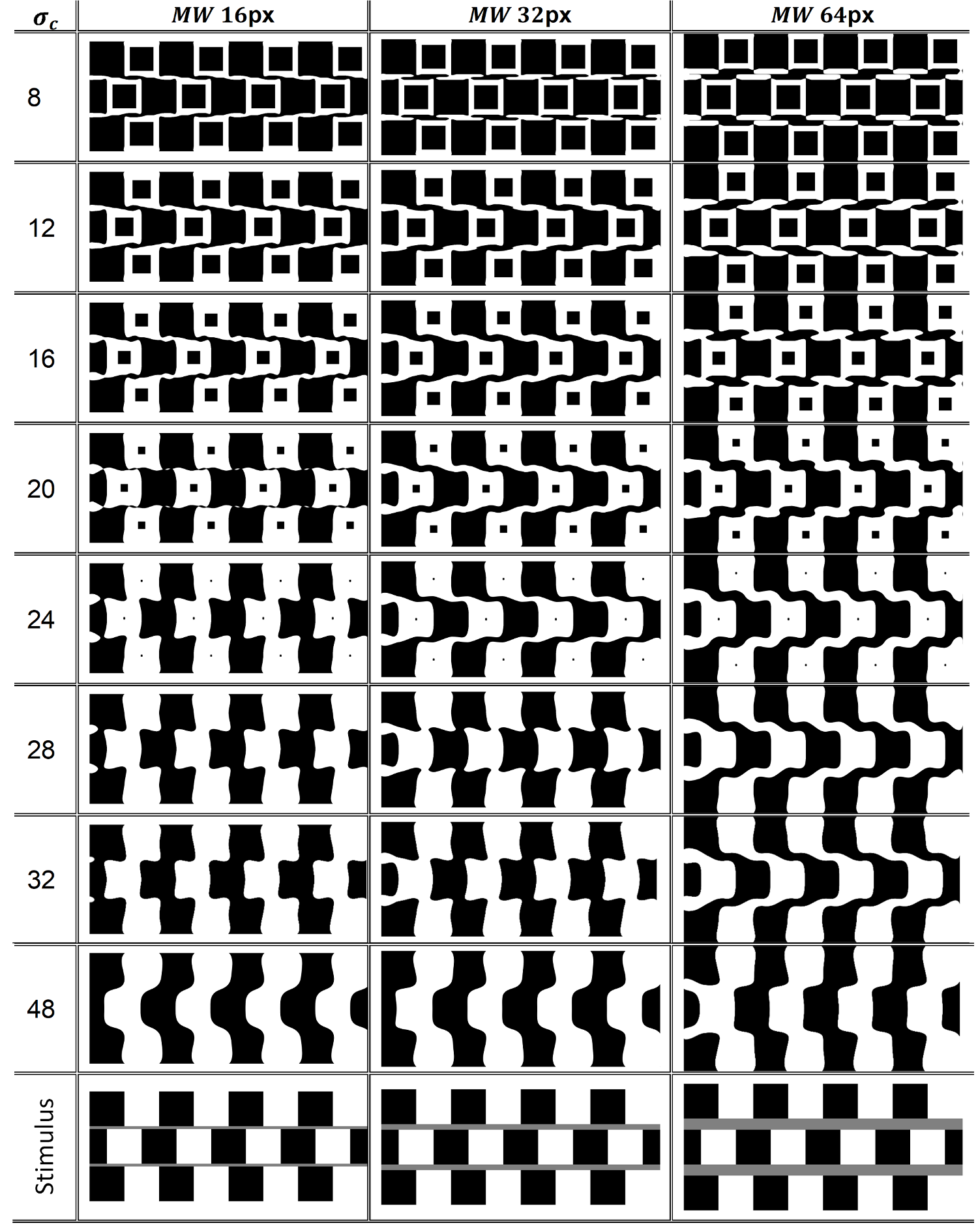}
  \caption{Binary DoG edge maps at multiple scales ($\sigma_c$) for three thick-mortar variations with mortar width of $MW=16$, 32 and 64px for the Caf\'e Walls of 3$\times$8 tiles with 200$\times$200px tiles. The other parameters of the DoG model are $s= 2$ (\emph{Surround ratio}) and $h= 8$ (\emph{Window ratio}). Reproduced by permission from [25].}
  \label{fig:B4}   
\end{figure*}
\begin{figure*}[htp]
  \centering
  \includegraphics[width=0.65\textwidth]{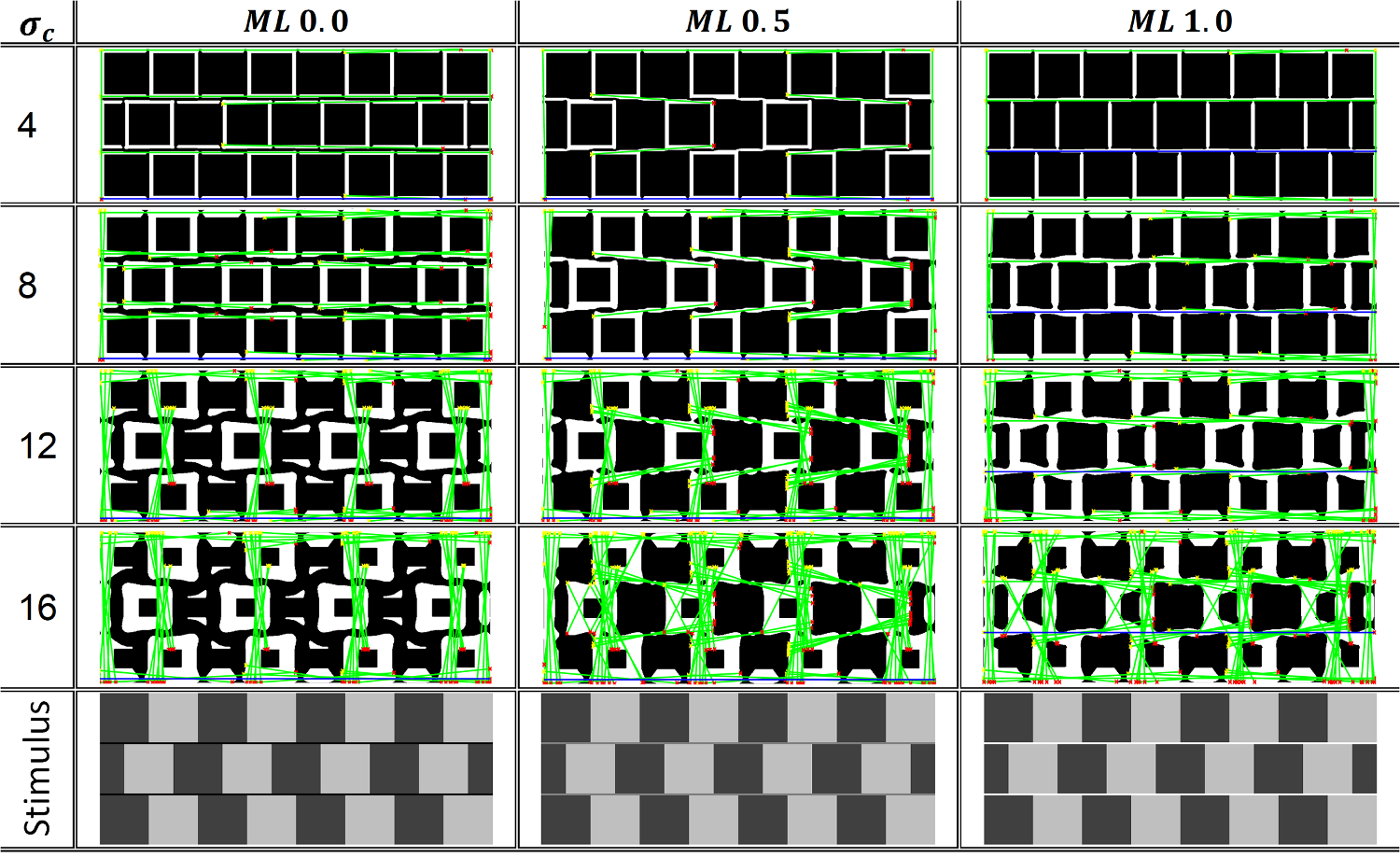}
  \caption{DoG edge maps at fine to medium scales ($\sigma_c$) for three Grey-Tiles variations with relative luminances of 0.25-Dark Grey, and 0.75-Light Grey, with the overlayed \textsf{Hough lines} displayed in green. Other parameters of the model and Hough processing stage are the same as Fig. B1.}
  \label{fig:B5}   
\end{figure*}
\begin{figure*}[htp]
  \centering
  \includegraphics[width=0.65\textwidth]{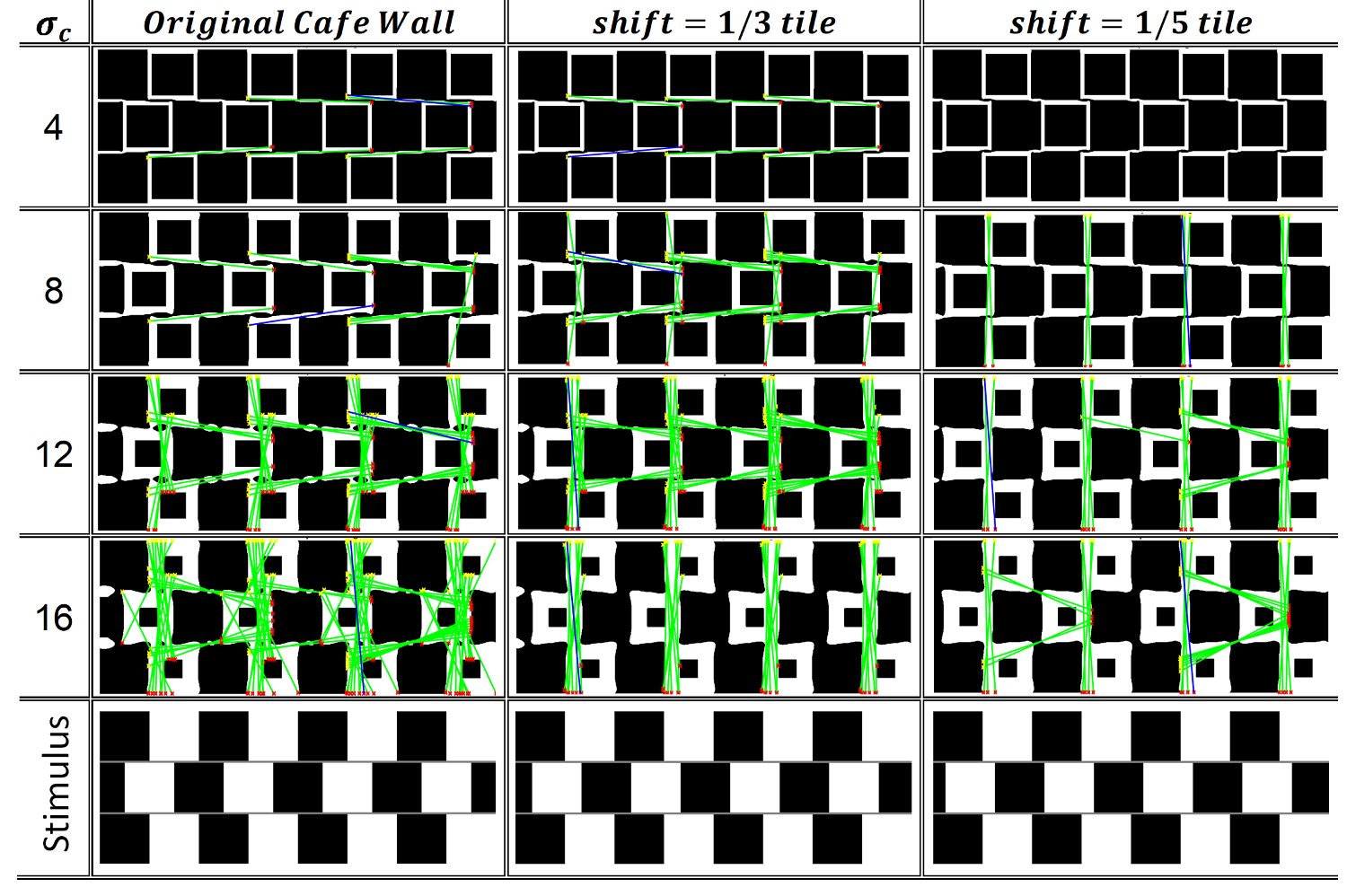}
  \caption{DoG edge maps at fine to medium scales ($\sigma_c$) for three patterns of the original Caf\'e Wall and two variations of phase of tile displacements with the shifts of 1/3 and 1/5 of a tile, with the overlayed \textsf{Hough lines} displayed in green. Other parameters of the model and Hough processing stage are the same as Fig. B1.}
  \label{fig:B6}   
\end{figure*}
\begin{figure*}[htp]
  \centering
  \includegraphics[width=0.65\textwidth]{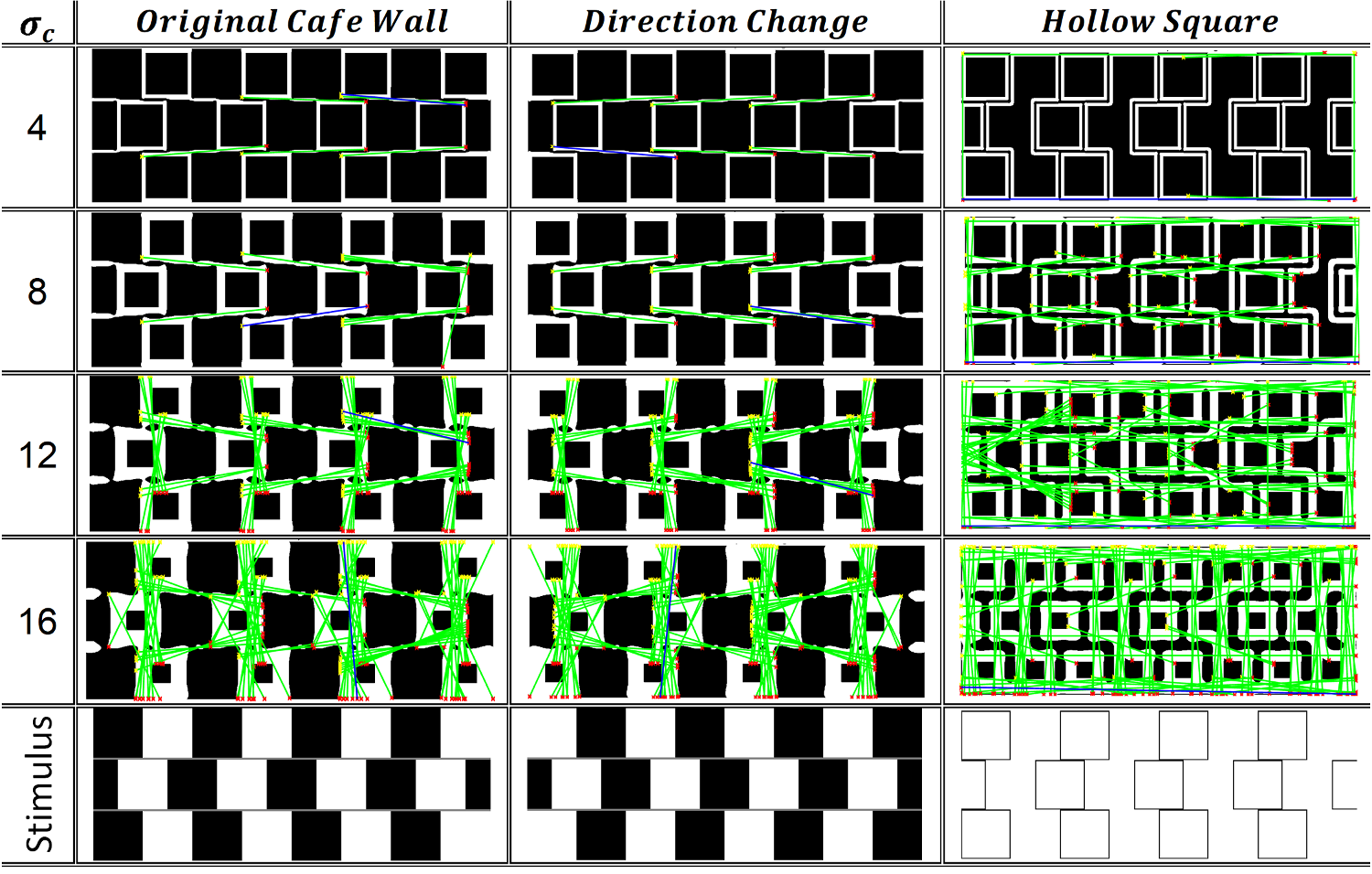}
  \caption{DoG edge maps at fine to medium scales ($\sigma_c$) for three patterns of the original Caf\'e Wall, the \emph{mirrored image} (\emph{Direction Change}) and the \emph{Hollow Square}, with the overlayed \textsf{Hough lines} displayed in green. Other parameters of the model and Hough processing stage are the same as Fig. B1.}
  \label{fig:B7}   
\end{figure*}
The DoG edge maps at multiple scales are presented here for all the variations tested in this study, in which the edge maps have been overlayed by the detected \textsf{Hough lines} displayed in green. Blue lines indicate the longest lines detected (see \hyperref[appendix:AppxA]{Appendix A} for the Hough analysis). The edge maps are shown at 7 scales for the `Mortar-Luminance' variations in \hyperref[fig:B1]{Fig. \ref*{fig:B1}}. The range of DoG scales starts from 0.5\emph{M}= 4px and continues to 3.5\emph{M}= 28px with incremental steps of 0.5\emph{M} ($\sigma_c= $4, 8, 12, 16, 20, 24, 28, \hyperref[sec:2.2]{Section \ref*{sec:2.2}}). Note that for the detection of near horizontal tilted line segments, as the perceived mortar lines in the Caf\'e Wall illusion, the DoG scale should be close to the mortar size, $\sigma_c\sim M= 8$px [20,21]. 

The difference between the DoG edge maps in between all the investigated patterns are mainly at fine to medium scales of the DoGs ($\sigma_c= $4, 8, 12, 16). As shown in \hyperref[fig:B1]{Fig. \ref*{fig:B1}} at medium scales ($\sigma_c= $12, 16), we see a transient state between detecting near horizontal tilted line segments connecting tiles with the Twisted Cord elements along the mortar lines, to zigzag vertical grouping of tiles at the coarse scales, in a completely opposite direction. The edge maps at coarse scales ($\sigma_c= $20, 24, 28) show very similar DoG outputs. This is because at the coarse scales, the scale of the DoG is large enough to capture the tile information. Therefore for the rest of the investigated patterns, we just show the edge maps \emph{at fine to medium scales} with the overlayed \textsf{Hough lines}. The edge maps for the `Mortar-Width' variations are presented in \hyperref[fig:B2]{Figs \ref*{fig:B2}} and \hyperref[fig:B3]{ \ref*{fig:B3}} as well as in the extended version of the DoGs for the very thick mortar lines in \hyperref[fig:B4]{Fig. \ref*{fig:B4}} to cover up not just the illusory tilts along the mortar lines, but also the groupings of tiles in zigzag vertical orientation at coarse scales in the thick mortar variations. For the rest of the stimuli tested, noted as `Significant Variations' in \hyperref[fig:4]{Fig. \ref*{fig:4}}, the edge maps at fine to medium scales with the overlayed \textsf{Hough lines} are shown in \hyperref[fig:B5]{Figs \ref*{fig:B5}} to \hyperref[fig:B7]{\ref*{fig:B7}}. We show the mean tilt angles measured in the DoG edge maps across multiple scales in \hyperref[appendix:AppxC]{Appendix C}.

\renewcommand{\theHsection}{C\arabic{section}}
\setcounter{secnumdepth}{0}
\counterwithin{figure}{section}
\newcommand{\hbAppendixPrefixC}{C}
\renewcommand{\thefigure}{\hbAppendixPrefixC\arabic{figure}}
\setcounter{figure}{0}
\section{\\C. Quantitative mean tilts}
\label{appendix:AppxC}
\begin{figure*}
  \centering
  \includegraphics[width=\textwidth]{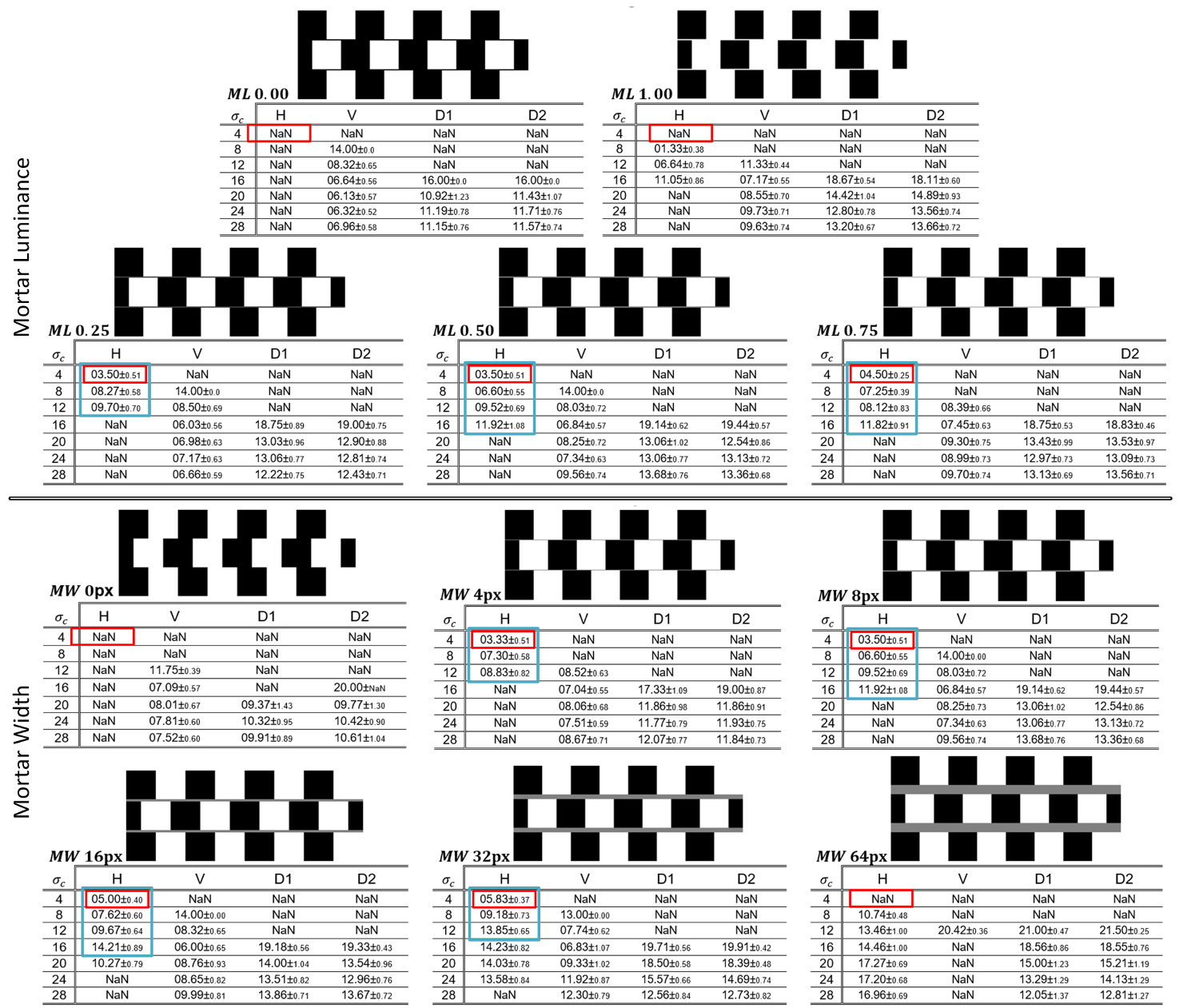}
  \caption{Mean tilts and the standard errors of detected tilt angles of \textsf{Hough lines} for (Top): Mortar-Luminance variations from Black (\emph{ML} 0.00) and White (\emph{ML} 1.00) mortars on Top, to Grey mortar lines ($ML=0.25$, 0.50, 0.75) for DoG edge maps at seven scales ($\sigma_c$). (Bottom): Mortar-Width variations from $MW=0$ to 64px. NaN: not a number means no lines are detected (Hough parameters are kept the same in all experiments; \hyperref[appendix:AppxA]{Appendix A}). We have highlighted the `foveal tilt effects' (FTE) in `red' and the `range of detected mean tilts' at fine to medium scales in `blue' boxes for the variations tested. Blue boxes also indicate the corresponding values for the `persistence of mortar cues' (PMC) in these variations (Please check \hyperref[sec:3.1.2]{Section \ref*{sec:3.1.2}} and \hyperref[fig:6]{Fig. \ref*{fig:6}} for more information). If there are no lines detected at scale 4, the PMC feature is set to \emph{None} and the FTE to $0.0^{\circ}$. If the PMC last till scale 8, we set the PMC to \emph{ML}. For Caf\'e Wall patterns with the PMC reaching scale 12, the PMC is set to \emph{M}. When the PMC reaches scale 16, we set the PMC to \emph{MH}, and above this scale ($\sigma_c>16$), the PMC value is set to \emph{H}. These features are used to predict the tilt effect in variations of the Caf\'e Walls tested.}
  \label{fig:C1}   
\end{figure*}

\begin{figure*}
  \centering
  \includegraphics[width=\textwidth]{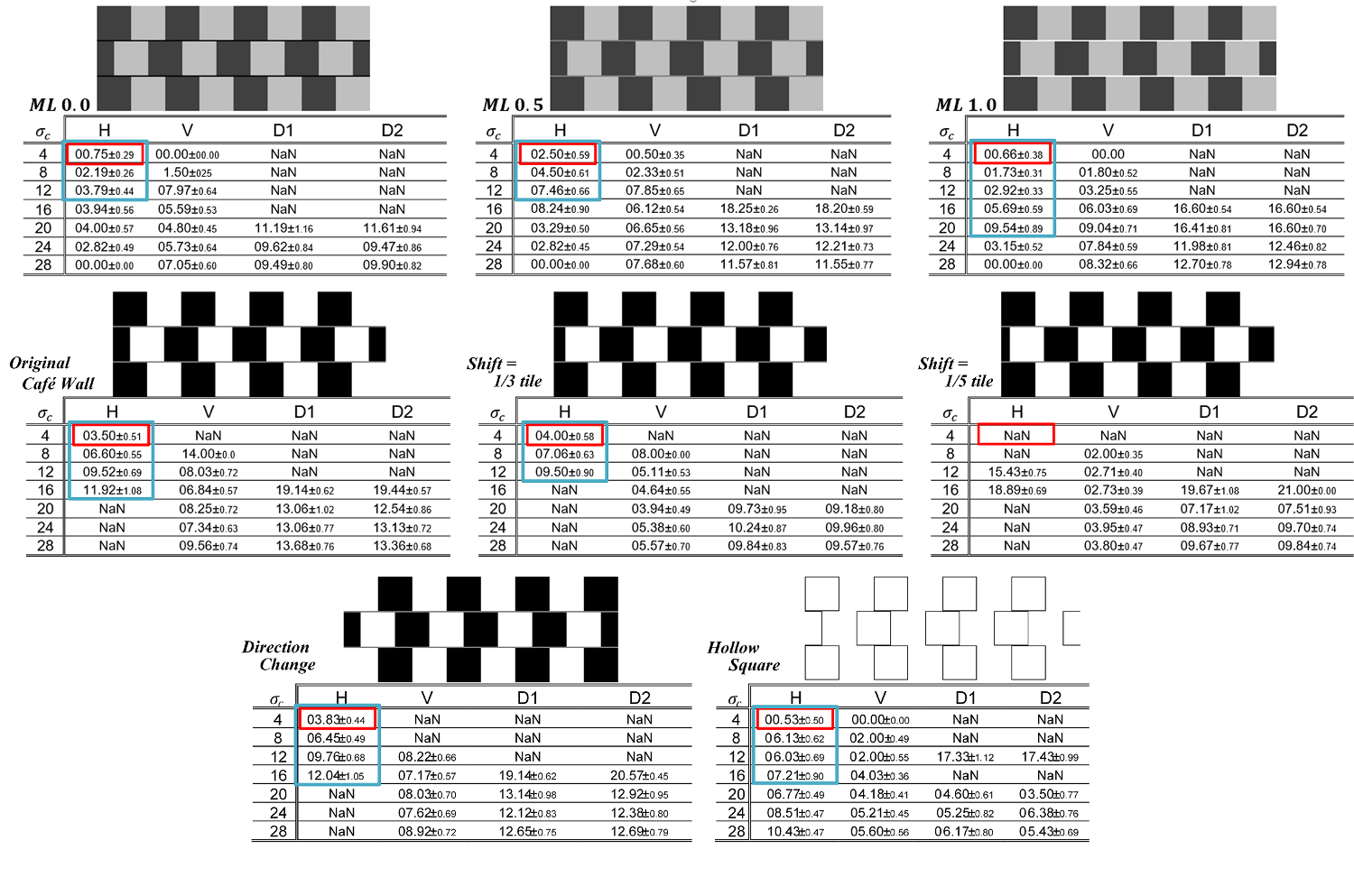}
  \caption{Mean tilts and the standard errors of detected tilt angles of \textsf{Hough lines} for (Top): Three variations of Grey-Tiles, (Middle): Three patterns of the original Caf\'e Wall pattern and two variations of phase of tile displacements with the shifts of 1/3 and 1/5 tile, (Bottom): The mirrored image and the Hollow Square version. The calculations are done for the edge maps at seven scales ($\sigma_c$) with fixed Hough parameters (\hyperref[appendix:AppxA]{Appendix A}) to detect near horizontal, vertical and diagonal tilted lines (H, V, D1, D2) in the edge maps (NaN means no lines are detected). For more information on the two features of FTE and PMC that are derived from these mean tilt tables please refer to \hyperref[fig:C1]{Fig. \ref*{fig:C1}} and the information provided in \hyperref[sec:3.1.2]{Section \ref*{sec:3.1.2}}.}
  \label{fig:C2}   
\end{figure*}

The absolute mean tilts and the standard errors of detected tilt angles for the Caf\'e Wall variations tested are provided here in Figs \hyperref[fig:]{Figs \ref*{fig:C1}} and \hyperref[fig:C2]{\ref*{fig:C2}}. For the `foveal tilt effect' (\emph{FTE}, explained in \hyperref[sec:3.1.1]{Sections \ref*{sec:3.1.1}} and \hyperref[sec:3.1.2]{Section \ref*{sec:3.1.2}}), we used the near horizontal mean tilts at scale 4, and reflected these values to \hyperref[fig:6]{Fig. \ref*{fig:6}} (in White) for all the variations tested. The FTEs are highlighted within `red boxes' in \hyperref[fig:C1]{Figs \ref*{fig:C1}}, \hyperref[fig:C2]{\ref*{fig:C2}}. We used another feature for our evaluations of tilt in the Caf\'e Wall illusion, called ``persistence of mortar cues'' (\emph{PMC}), that indicates the strength of the tilt effect in variations of the Caf\'e Wall illusion tested. Please check \hyperref[sec:3.1.2]{Section \ref*{sec:3.1.2}} to find out how to evaluate the PMC values corresponding to each stimulus tested. The values for this feature include \emph{None, ML:Medium Low, L:Low, M:Medium, MH:Medium High, and H:High} as reflected in \hyperref[fig:6]{Fig. \ref*{fig:6}} (in Blue). This feature provides a qualitative metric for us showing the strength of the tilt effect in this illusion. The PMC value can be extracted from the corresponding scales next to the highlighted blue boxes in \hyperref[fig:C1]{Figs \ref*{fig:C1}}, \hyperref[fig:C2]{\ref*{fig:C2}}. For a complete analysis of the tilt effect in the Caf\'e Wall, we need these two features (\emph{FTE, PMC}) as well as the `range of detected mean tilts' as highlighted by the blue boxes. Please check \hyperref[sec:3.4]{Section \ref*{sec:3.4}} and \hyperref[fig:8]{Fig. \ref*{fig:8}} for the final outcome of our tilt investigations for all the Caf\'e Wall stimuli tested in this study.
\end{subappendices}
\begin{acknowledgements}
Nasim Nematzadeh was supported by an Australian Research Training Program (RTP) award. The authors are grateful to Mrs. Sue Powers and Prof. Shanbehzadeh for their critical reading and careful comments on a previous draft of this paper. The authors declare that there is no conflict of interest regarding the publication of this paper.
\end{acknowledgements}



\end{document}